\documentclass[letterpaper, 10pt, twocolumn]{article}
\usepackage{clean}
\shorttitle{Simulation for optimization of compliant fingers}
\AddToHook{env/tabular/begin}{\tiny}

\usepackage{textcomp, gensymb}

\usepackage{adjustbox}
\usepackage{graphicx}
\usepackage{subfig}
\graphicspath{{./figs}}
\usepackage{epsfig}

\usepackage{amsmath,amssymb,amsfonts}
\usepackage{algorithm2e}
\usepackage{xcolor}
\usepackage[colorlinks=true, linkcolor=black, urlcolor=cyan, filecolor=black, citecolor=black]{hyperref}
\usepackage{url}
\usepackage{amsmath,bm}
\usepackage{multirow, makecell}
\usepackage{adjustbox}
\usepackage{comment}
\def\code#1{\texttt{#1}}
\let\labelindent\relax
\usepackage{tikz}
\usetikzlibrary{tikzmark}

\usepackage{float}
\usepackage{enumitem}
\usepackage{booktabs}
\usepackage{cite} % bibtex, should be used for IEEE

\begin{document}
% Sim2real gap of compliant fingers in high-speed assembly 
% Towards simulation-based optimization of compliant fingers for high-speed connector assembly
% \title{Structured compliance design optimization by dynamic simulation}
\title{Towards simulation-based optimization of compliant fingers \\ for high-speed connector assembly}

\author{Richard Matthias Hartisch$^1$, Alexander Rother, Jörg Krüger$^{1,2}$, Kevin Haninger$^2$} 

\maketitle

\footnotetext[1]{Department of Industrial Automation Technology at TU Berlin, Germany.}
\footnotetext[2]{Department of Automation at Fraunhofer IPK, Berlin, Germany.}

%With corresponding author: 

\renewcommand\thefootnote{}
\footnotetext{Corresponding author: \texttt{r.hartisch@tu-berlin.de}. This project has received funding from the European Union's Horizon 2020 research and innovation programme under grant agreement No.\ 101058521 — CONVERGING.}
\renewcommand\thefootnote{\arabic{footnote}}

%

% Without: 

% \renewcommand\thefootnote{}
% \footnotetext{This project has received funding from the European Union's Horizon 2020 research and innovation programme under grant agreement No.\ 101058521 — CONVERGING.}
% \renewcommand\thefootnote{\arabic{footnote}} 

%%% END ORIGINAL 
% \thanks{\noindent$^1$ Department of Industrial Automation Technology at TU Berlin, Germany. \\ $^2$ Department of Automation at Fraunhofer IPK, Berlin, Germany.  \\ Corresponding author: {\tt r.hartisch@tu-berlin.de}}
% \thanks{This project has received funding from the European Union's Horizon 2020 research and innovation programme under grant agreement No 101058521 — CONVERGING.}}

% \footnotetext{Corresponding author: \texttt{r.hartisch@tu-berlin.de} This project has received funding from the European Union's Horizon 2020 research and innovation programme under grant agreement No 101058521 — CONVERGING.}

% \begin{center}

% $^1$ Department of Industrial Automation Technology at TU Berlin, Germany. \\
% $^2$ Department of Automation at Fraunhofer IPK, Berlin, Germany. \\
% Corresponding author: \texttt{r.hartisch@tu-berlin.de} \\
% This project has received funding from the European Union's Horizon 2020 research and innovation programme 
% under grant agreement No 101058521 — CONVERGING.
% \end{center}

% % \IEEEoverridecommandlockouts
% \maketitle

\begin{abstract}
% 4 parts to an abstract: 
%   1. what's happening today, 
%   2. what's missing or bad about that, 
%   3. what do you propose, 
%   4. how do you validate, what is improved.

Mechanical compliance is a key design parameter for dynamic contact-rich manipulation, affecting task success and safety robustness over contact geometry variation.
Design of soft robotic structures, such as compliant fingers, requires choosing design parameters which affect geometry and stiffness, and therefore manipulation performance and robustness. 
Today, these parameters are chosen through either hardware iteration, which takes significant development time, or simplified models (e.g. planar), which can't address complex manipulation task objectives. Improvements in dynamic simulation, especially with contact and friction modeling, present a potential design tool for mechanical compliance. We propose a simulation-based design tool for compliant mechanisms which allows design with respect to task-level objectives, such as success rate. This is applied to optimize design parameters of a structured compliant finger to reduce failure cases inside a tolerance window in insertion tasks. The improvement in robustness is then validated on a real robot using tasks from the benchmark NIST task board. The finger stiffness affects the tolerance window: optimized parameters can increase tolerable ranges by a factor of $2.29$, with workpiece variation up to $8.6 \ mm $ being compensated. However, the trends remain task-specific. In some tasks, the highest stiffness yields the widest tolerable range, whereas in others the opposite is observed, motivating need for design tools which can consider application-specific geometry and dynamics.

\end{abstract}

\section{Introduction} \label{sec:intro}

Robust contact-rich manipulation requires compensation for small errors in the contact geometry \cite{Hartisch_TMECH, li2019survey}. 
% Dynamic contact-rich manipulation requires this be done with high bandwidth, such that the compensation performs well even at higher speeds \cite{wang1998passive}. 
Mechanical compliance offers compensation of variation in contact geometry through intrinsic dynamics; reducing the sensitivity to position variations and enabling self-correcting intrinsic dynamics, e.g. self-centering \cite{ciblak2003design, yun2008compliant}. The importance of compliance can be seen from its robotic applications: integrated in the kinematic structure \cite{herron2024pandora}, joints \cite{albu2008soft}, flange \cite{ciblak2003design, kim2021displacement}, fingers \cite{shintake2018soft, Hartisch_TMECH}, and environment \cite{bhagat2014design, hartisch2022flexure, haninger2022contact}.

Mechanical compliance directly affects normal forces and surface geometry during contact. This directly affects pressure distribution on contact areas, Coulomb friction limits, and the robustness uniformity over variation in object contact condition.

The mechanical design of compliance, however, is limited to lower-dimensional, simplified models (e.g. planar systems) to enable design analysis such as constraint-based design \cite{su2009screw}, effective TCP stiffness matrix \cite{ang2002specifying, hartisch2022flexure}, or compliance ellipsoids \cite{kim2008building}. While these methods can develop design intuition, they cannot scale to more complex designs. Topological optimization methods can improve the range of motion or linearity of compliant mechanisms \cite{alacoque2025compliant}, but has not been investigated for task-related performance. 
% Topological optimization methods can improve the range of motion or linearity of compliant mechanisms \cite{alacoque2025compliant}, but not task-related performance. The geometry of a finger can be optimized 

% Control parameters can be more easily optimized from trial-and-error as they do not require hardware iterations. For example, the virtual stiffness of an impedance controller can be optimized \cite{oikawa2021reinforcement, chang2022impedance}. However, mechanical parameters do not have the bandwidth limitations of active control \cite{Hartisch_TMECH}, therefore methods for mechanical compliance design are considered here. As 3D printed compliant structures enables a larger design space, methods are needed to optimize these designs with respect to task-level performance: e.g. robustness over part variation, in-hand slip, maximum collision force.
Mechanical parameters do not have the bandwidth limitations of active control \cite{Hartisch_TMECH}, therefore methods for mechanical compliance design are considered here. As 3D printed compliant structures enables a larger design space, methods are needed to optimize these designs with respect to task-level performance: e.g. robustness over part variation, in-hand slip, maximum collision force.

\begin{figure}
    \centering
    \includegraphics[width=\columnwidth]{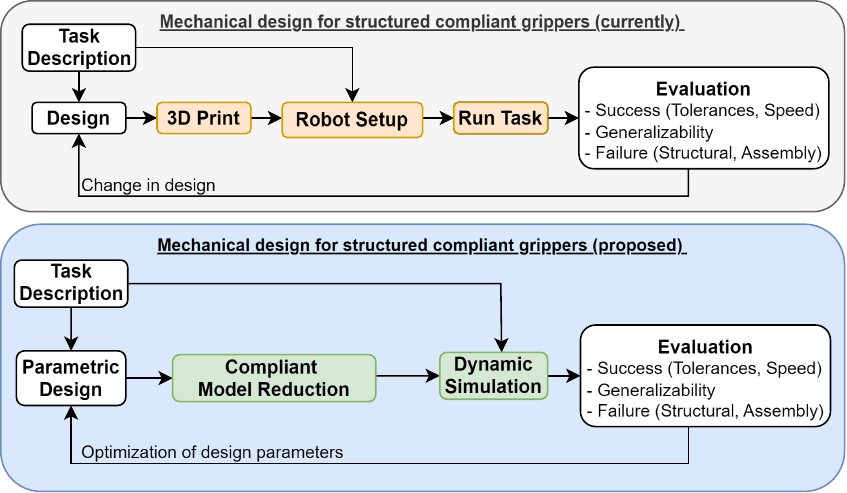} 
    \caption{Comparison of current (top) and proposed (bottom) design pipeline of passive, structured compliance. The objective is to reduce the need for real-life experiments, using instead a robot model and task (enter on the left) and dynamic simulation to evaluate a range of design and environmental parameters. The simulation results are evaluated according to robustness over part pose variation and failure cases. In this paper, these simulation results are compared against real execution.}
    \label{fig:dyn_sim_overview}

\end{figure}

At the same time, dynamic simulation has substantially improved for robotics in terms of geometric complexity it can handle, numerical stability of simulation, differentiability, and realism. This offers a potential design tool for compliance, one which can scale in task and design complexity as well as provide task-level evaluation. Soft systems, e.g. compliant grippers, are numerically easier to simulate compared to stiff systems, which may allow for a reduced sim2real gap \cite{acary2018solving}. However, the question of simulation accuracy remains critical: is simulation accurate enough to motivate the design decisions in compliant device design? Simulation has been used as a tool to iteratively optimize fingertip contact geometry and grasping success \cite{hagelskjaer2019combined}, but not yet applied for compliant fingers.

% While simulation has improved numerically, stiff systems are numerically difficult to integrate, requiring shorter integration time steps or application-specific tuning of parameters. Soft systems, e.g. compliant grippers, result in systems that are numerically easier to simulate. Compared to stiff solutions, as in rigid grippers, this may allow for a reduced sim2real gap \cite{acary2018solving}.

% Paragraph about simulation tools
% To answer this, we compared some of the most common simulation-tools for robotics to find one that is suited for this task. As there exist countless physics- engines, we narrowed down the selection by elaborating necessary requirements. The simulation-tool has to be simulate assembly tasks (higher number of contact points, in-hand slip, concave geometries) while accurately describing the tolerable ranges of the compliant grippers and assessing failure cases during the assembly phases.
% % with contacts and friction involved as well as be able to model a different stiffness in every spatial direction.
% While some physics-engines, e.g. PyBullet, NVIDIA Isaac and Google Brax, focus more on reinforcement learning and a GPU-accelerated simulation. In MuJoCo we found a physics-engine that focuses on contact-rich tasks with sufficient accuracy.

To investigate applicability, MuJoCo is chosen as a physics-engine that focuses on contact-rich tasks with sufficient accuracy, modeling the 6-DOF stiffness between object contact and robot flange, and investigating if the simulation can predict the tolerance window in a resulting assembly task. 
Additionally, the ability to determine failure cases, i.e. slip, a failed search strategy and jamming of the plug is investigated. An overview of the approach can be seen in Figure \ref{fig:dyn_sim_overview}.

\section{Task Description}
\label{sec:task}
This section describes the manipulation tasks considered, defining the assembly goal and failure cases. 

\subsection{Assembly Applications}

\begin{figure}[t]
\centering
\scalebox{.95}{ % adjust scale factor as needed
    \begin{tabular}{cc}
        \subfloat[KET12\label{fig:KET12}]{
            \includegraphics[width=0.45\columnwidth]{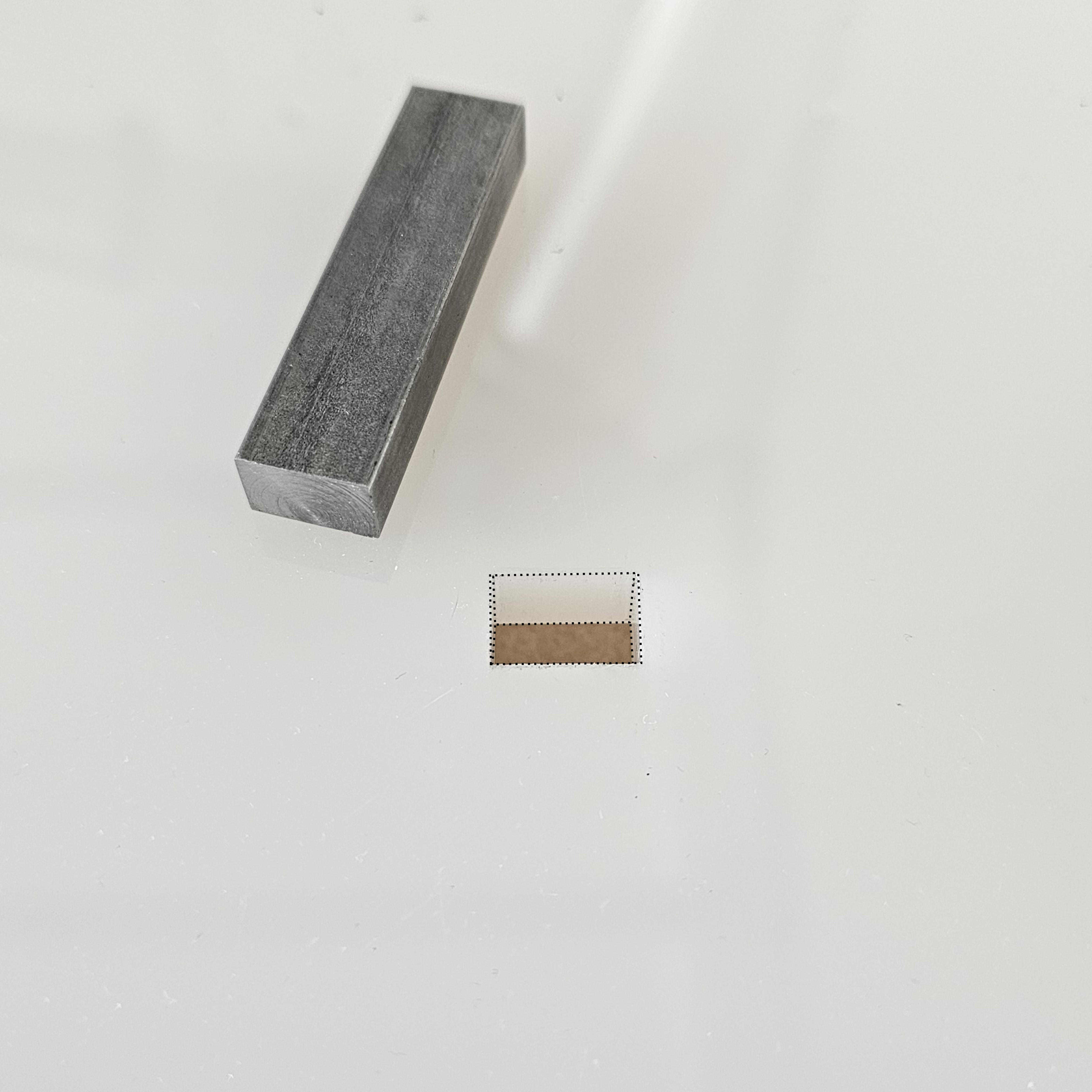}} &
        \subfloat[KET8\label{fig:KET8}]{
            \includegraphics[width=0.45\columnwidth]{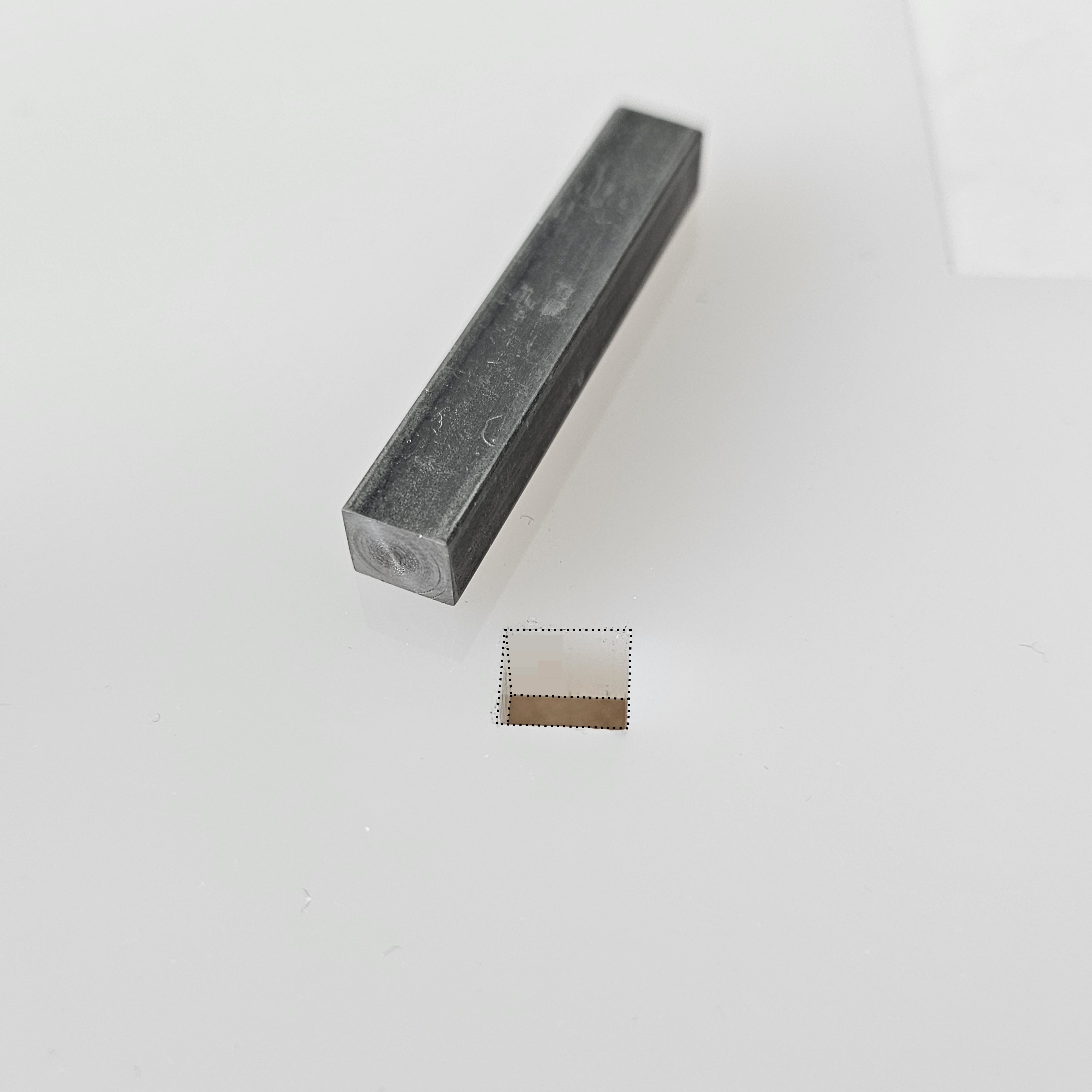}} \\[0.3cm]
        \subfloat[RJ45\label{fig:RJ45}]{
            \includegraphics[width=0.45\columnwidth]{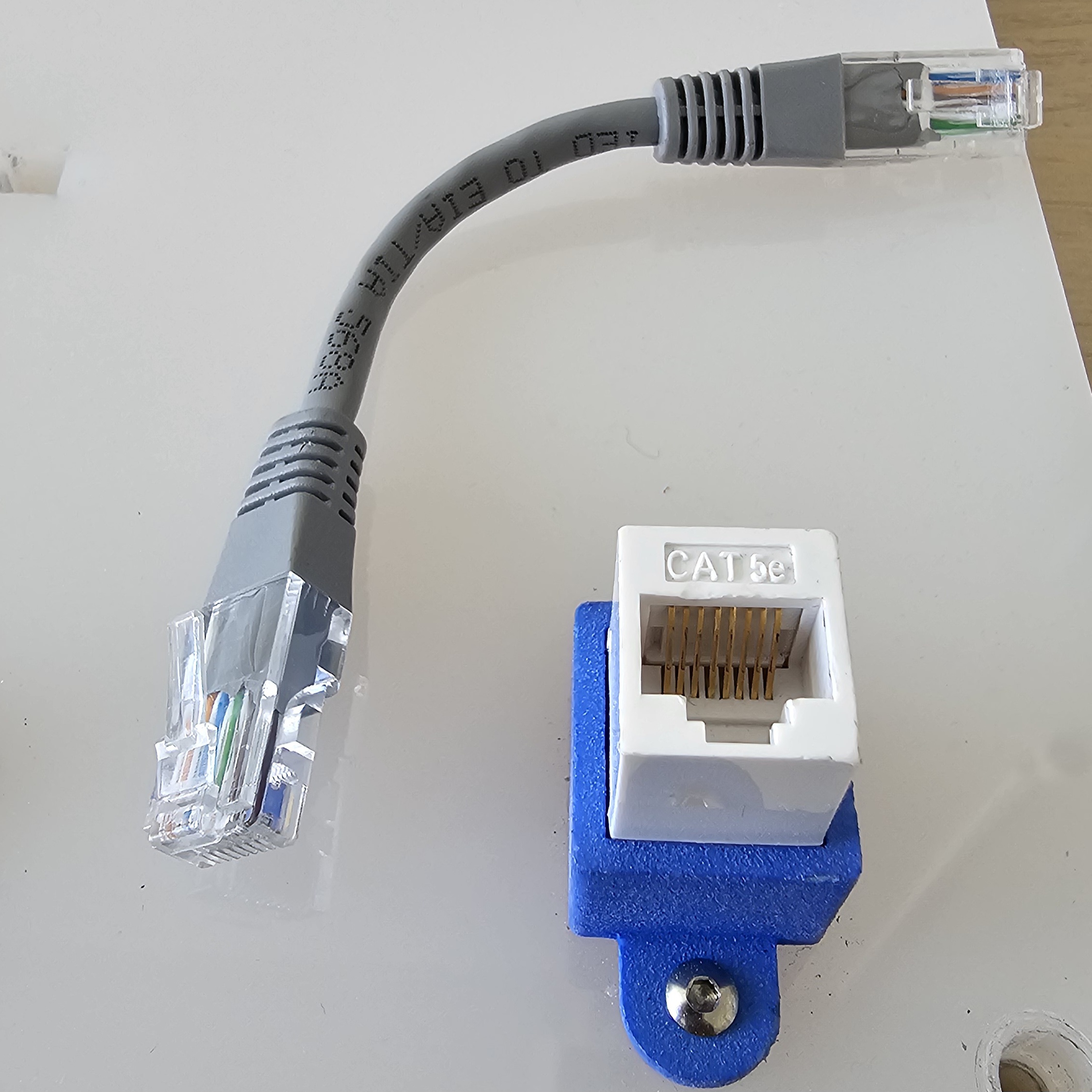}} &
        \subfloat[USB-A Connector\label{fig:USB}]{
            \includegraphics[width=0.45\columnwidth]{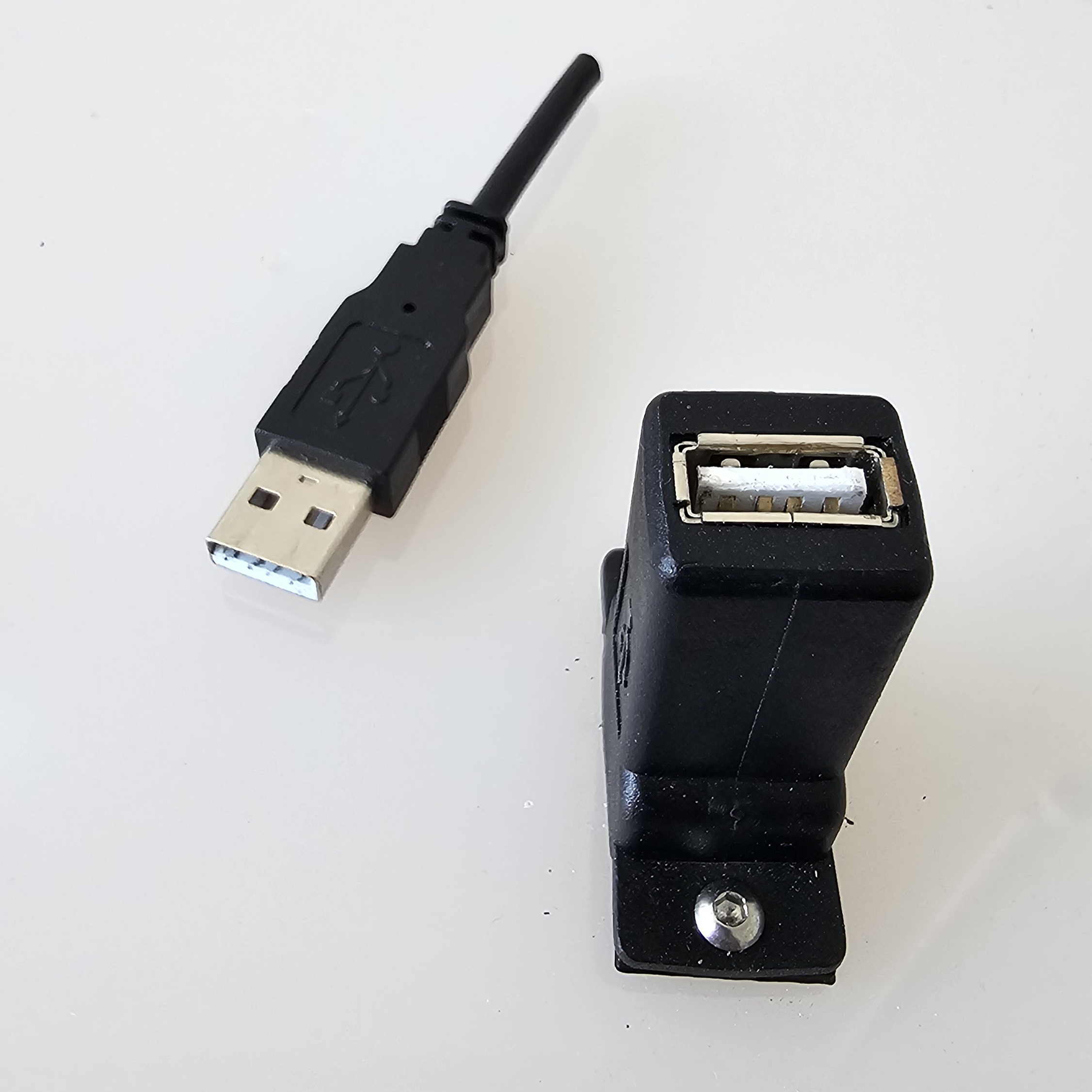}}
    \end{tabular}
}
\caption{The assembly tasks considered in this paper are shown.}
\label{fig:tasks}

\end{figure}

Four assembly tasks have been chosen from the NIST task board, which are both simulated and carried out in real-life experiments. These include two rectangular shaped pegs in varying sizes (KET12 in Fig. \ref{fig:KET12} and KET8 in Fig. \ref{fig:KET8}), as well as two electrical connectors (RJ45 in Fig. \ref{fig:RJ45} and USB in Fig. \ref{fig:USB}). The assembly sequences utilized are similar for all assembly tasks and are further described in the simulation preparation in Sec \ref{sec:movement}. 

\subsection{Robot Assembly Process}
\label{sec:assembly}

\begin{figure}[t]
\centering
\scalebox{.90}{ % adjust overall size here
    \begin{tabular}{ccc}
        \subfloat[\label{fig:usb angled}]{
            \includegraphics[width=0.3\columnwidth]{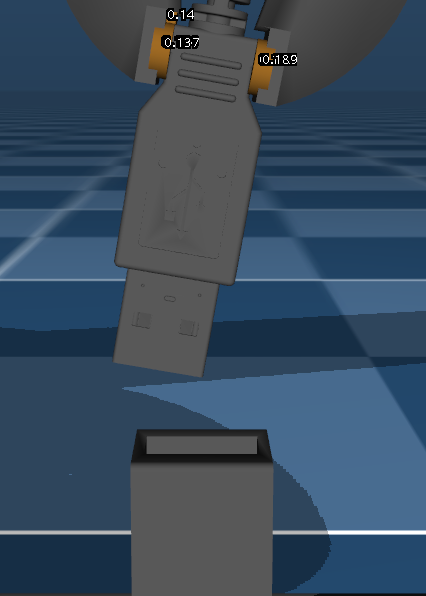}} &
        \subfloat[\label{fig:usb top}]{
            \includegraphics[width=0.3\columnwidth]{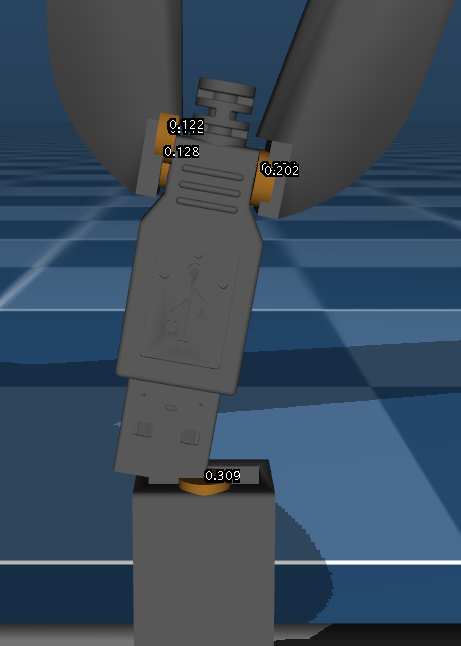}} &
        \subfloat[\label{fig:usb back}]{
            \includegraphics[width=0.3\columnwidth]{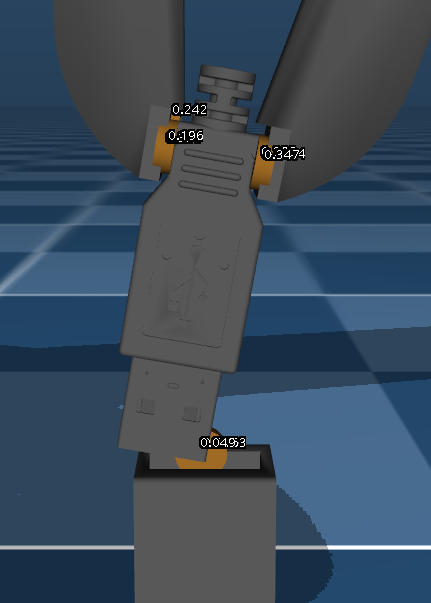}} \\[0.3cm]
        \multicolumn{3}{c}{
            \subfloat[\label{fig:usb front}]{
                \includegraphics[width=0.3\columnwidth]{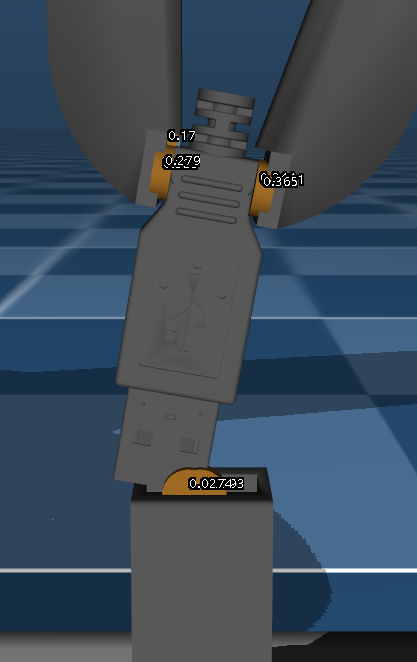}}
            \hspace{0.5cm} % adjust spacing between last two
            \subfloat[\label{fig:usb side}]{
                \includegraphics[width=0.3\columnwidth]{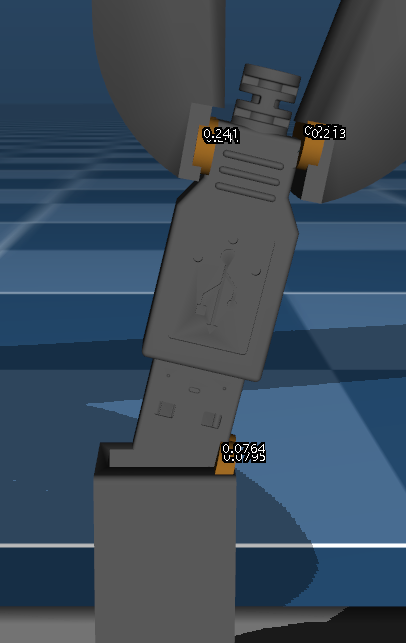}}
        }
    \end{tabular}
}
\caption{Here, the simulated assembly movement is demonstrated exemplary for the USB assembly task. (a) and (b) visualize the approaching movement and first established contact on the socket. In (c) and (d) a back-and-forth motion of the connector in the  x- direction with a slight movement in the z- direction is executed to allow the plug to slip into the socket when aligned. Following, in (e) contact is established between the sides of the plug and socket, realized by a motion in the negative y- direction.}
\label{fig:search}

\end{figure}

To achieve high-speed insertion, compliant fingers are used along with an assembly strategy, based on the search strategy employed in \cite{Hartisch_TMECH}, which can can be seen in Fig. \ref{fig:search}.
First, the robot is positioned in its home position. Next, the robot is moved from the home position to above the connector within a safe distance. Next, it is lowered to the gripping height, after which the grippers are closed, grasping the part. The robot then moves upwards again to the previous position, while holding on to the connector. The aim of this process is to ensure that the connector is aligned with the socket before the assembly is performed via search strategy. The process is identical for all assembly tasks, except the starting position and the distance that the robot has to move for alignment are different.

% \begin{figure} [t]
% \scalebox{.5}
%     \centering
%     \subfloat[
% 	\label{fig:usb angled}]{\includegraphics[width=.3\columnwidth]{figs/USB angled2.png}}
%     \hfill
%     \subfloat[
% 	\label{fig:usb top}]{\includegraphics[width=.3\columnwidth]{figs/USB top.png}}
% 	\hfill
%     \subfloat[
% 	\label{fig:usb back}]{\includegraphics[width=.3\columnwidth]{figs/USB back.png}}
%     \hfill
%     \subfloat[
% 	\label{fig:usb front}]{\includegraphics[width=.3\columnwidth]{figs/USB front.png}}
% 	\hspace{1cm}
%     \subfloat[
% 	\label{fig:usb side} ]{\includegraphics[width=.3\columnwidth]{figs/USB side.png}}
%     \caption{Here, the simulated assembly movement is demonstrated exemplary for the USB assembly task. (a) and (b) visualize the approaching movement and first established contact on the socket. In (c) and (d) a back-and-forth motion of the connector in the  x- direction with a slight movement in the z- direction is executed to allow the plug to slip into the socket when aligned. Following, in (e) contact is established between the sides of the plug and socket, realized by a motion in the negative y- direction.}
%     % \vspace{-0.5cm}
%     \label{fig:search}
% \end{figure}

\subsection{Failure Cases}

Practical design optimization for robotic tasks means reducing the incidence of failure cases, which we define next. Scenarios in which the connector is not properly assembled in the socket are defined as an unsuccessful assembly task. 

\begin{figure}[t] 
    
	\centering
	\subfloat[Failure Case 2) - Plug missed
	\label{fig:missed}]{\includegraphics[width=.30\columnwidth]{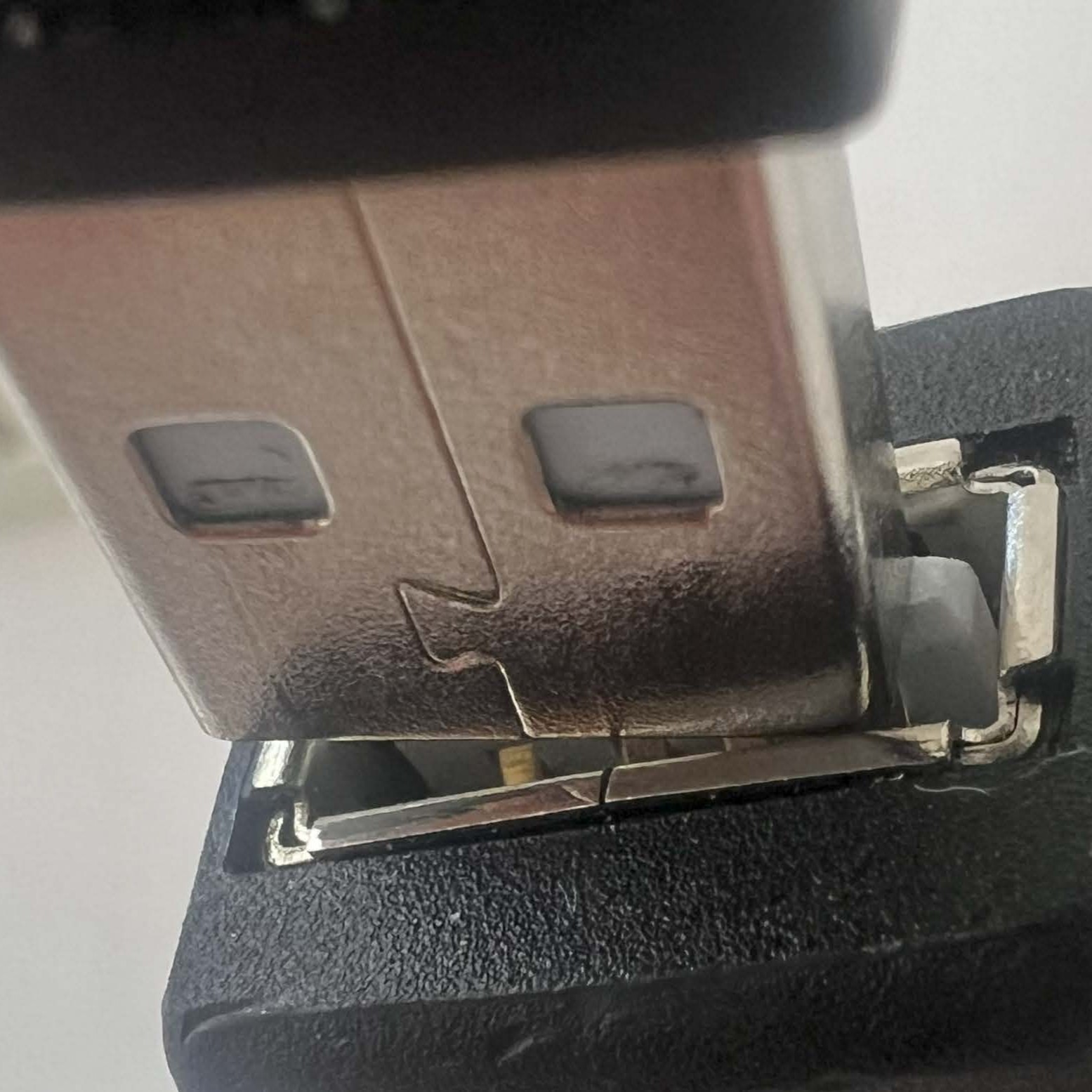}}
	\hfill
	\subfloat[Failure Case 3) - Connector jammed
	\label{fig:jammed}]{\includegraphics[width=.30\columnwidth]{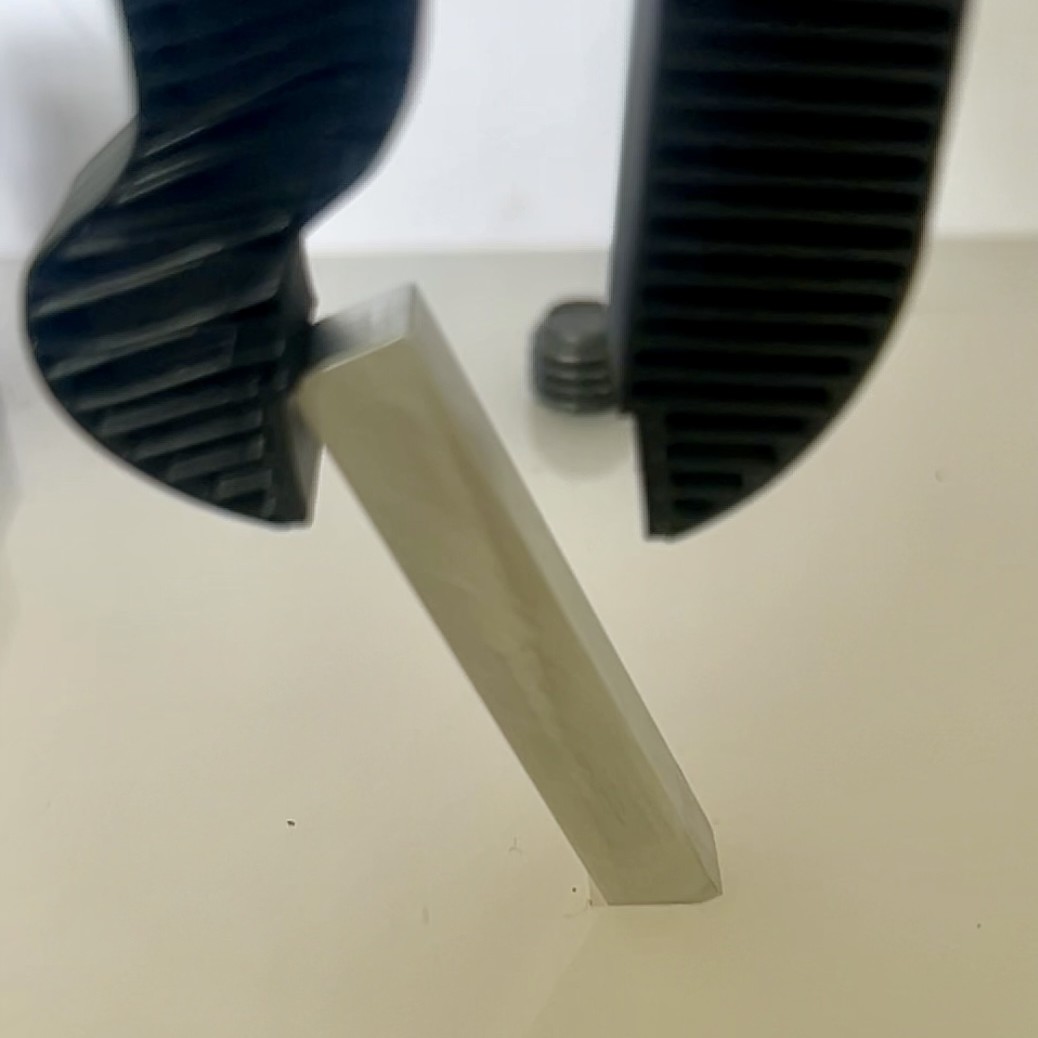}}
	\hfill
	\subfloat[Failure Case 4) - Contact loss
	\label{fig:loss}]{\includegraphics[width=.30\columnwidth]{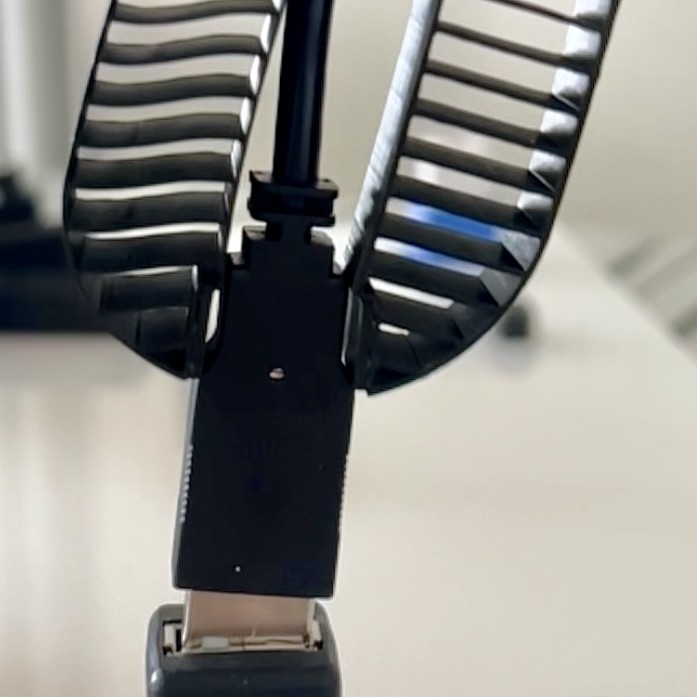}}
	\hfill
	\protect\caption{Visualized Failure Cases}
	\label{fig:failurecases}

\end{figure}

During the assembly process, following, often interconnected, failure cases are possible and have been noted, which can be subdivided into four error sources: failure of the grippers, as demonstrated in Fig. \ref{fig:failurecases}, failed assembly due to the robot movement, occurring external forces and error sources from the plug and socket geometry and dynamics. The individual observed failures within the occurring assembly phases and the corresponding failure causes are listed in Tab. \ref{tab:failurecasesrev}in correspondence to the phase of assembly they are observed in. Failure case 1) depicts contact loss during grasping, manipulation and approaching, which has been noted by parameter combinations of low infill density and stiffness in combination with soft materials for PETG grippers \cite{Hartisch_TMECH}. 
Failure case 2) emphasizes on a failed search phase in which the plug is missed, as seen in Fig. \ref{fig:missed}, either due to a deficient search strategy movement and/or contact with the socket geometry. Both during the alignment and insertion the connector can be jammed, which is described in failure case 3) and visualized in Fig. \ref{fig:jammed}. This occurs either due to insufficient stiffness in the assembly direction, occurring external forces from both dynamics and geometry of the connector, i.e. chamfers, which prevent centering of the connectors and therefore result in collision and jamming. Contact loss as failure case 4) is derived from insufficient normal- and friction force of the gripper and excessive external forces during contact between connector and socket, resulting in a loss of grip. 

% While the subdivision of the failure cases in the assembly phases and failure causes sounds intuitive, it has to be noted, that these failure cases are often interconnected and have overlapping failure causes, which complicates a clear division of these. 

\begin{table*} [t]
\caption{Failure Cases}
\centering
\resizebox{\textwidth}{!}{
\begin{tabular}{| l | l || c c c | c | c | c c | c |} 
\hline

\multirow{3}{*}{\textbf{Assembly Phase}} & \multirow{3}{*}{\textbf{Observed Problem}} &  \multicolumn{8}{c|}{\textbf{Failure Causes}} \\

\cline{3-10}

 &  & \multicolumn{3}{c |}{\textbf{Gripper}} & \textbf{Movement} &  \multirow{2}{*}{\textbf{External Forces}} &  \multicolumn{2}{c|}{\textbf{Plug}} & \textbf{Socket}\\
\cline{3-6}
\cline{8-10}

&  & Normal Force & Friction Force & Insufficient Stiffness & Search Strategy  & & Dynamics & Geometry &  Geometry \\

\hline
\hline
% &&& \\[-1.5ex]

Grasping and Manipulation, &  \multirow{2}{*}{1) Contact loss \label{failure:contactloss}} & \multirow{2}{*}{X} & \multirow{2}{*}{X}   & \multirow{2}{*}{X}  &  &  &  & & \\

Approaching &  &  &  &  &  &  &  & & \\
\hline
% &&& \\[-1.5ex]
Searching  & 2) Plug missed \label{failure:plugmissed} &    &     &     & X &  &  & & X \\
\hline
% &&& \\[-1.5ex]
% Alignment  & 3) Connector jammed \label{failure:connectorjammedalignment} &  &  & X & X & X &  X & X & \\
% \hline
% % &&& \\[-1.5ex]

\multirow{ 2}{*}{Insertion}  & 3) Connector jammed \label{failure:connectorjammedinsertion}&  &  & X & X & X & X & X & \\
% \hline
% &&& \\[-1.5ex]
\cline{2-10}
% &&& \\[-1.5ex]
                              & 4) Contact loss \label{failure:contactlossinsert}  & X  &  X  & X &  & X &  & & \\

\hline
\end{tabular}}

\label{tab:failurecasesrev}
\end{table*}

\section{Modeling of Compliant Fingers}

% In this section, the modeling and parametrization of a structured compliant gripper, as proposed in \cite{Hartisch_TMECH} is described. 

% Soft fingers are usually designed to achieve universal gripping of surfaces with various geometries, which are either silicone-based or finray-based \cite{hao_universal_2016}, \cite{Liu.2020}, \cite{Park.2019old},\cite{crooks2016fin}, \cite{Elgeneidy.2019}. These classical grippers utilize an enveloping grip where deformation adapts to variation in surface geometry of grasped parts. 

The proposed design principle in \cite{Hartisch_TMECH} realizes a structured compliance through the internal structures of the fingers for passive alignment. Structured compliance allows the fingers to deflect under load, while providing sufficient stiffness, and therefore force, in the assembly direction, establishing contact only at the fingertip, as opposed to typical soft silicone- or finray-based soft grippers, which utilize an enveloping grip adapting to variation in surface geometry \cite{hao_universal_2016}, \cite{Liu.2020}, \cite{Park.2019old},\cite{crooks2016fin}, \cite{Elgeneidy.2019}. 
% The rendered compliance on a gripped part supports mechanical search and alignment, while providing sufficient force in the assembly direction. 

% In the following, Sec. \ref{sec:params} introduces the design parameters and the design principle of the compliant fingers. In Sec. \ref{sec:stiffness}, the stiffness model is described. 

\subsection{Design principle and parameters}
\label{sec:params}

% \begin{figure}[t]
%     \centering
%     \includegraphics[width=0.8\columnwidth]{figs/Finger_shematic.png} 
%     \caption{(a) gives an overview of the design parameters in the structured compliant gripper. Here, the notched fingertip, the infill density, the infill direction, as well as the mounting angle are visualized. Additionally, the compliance direction and the assembly direction are shown. In (b) the KET assembly task is visualized}
%     \label{fig:schematic}
% \end{figure}

The identified design parameters in \cite{Hartisch_TMECH} are the infill options to adjust the density and orientation of the ribs in the finger, i.e. the infill direction (i), given in $\degree$, and the infill density (ii), given in $ \%$. This affects the bulk stiffness realized by the finger on a gripped part, as well as the maximum force that the finger can apply. Additionally, the compliant structure of the fingers enables an effective remote center of compliance (RCC), such that a lateral translation at the base of the plug causes a change in orientation, with the tip moving in the direction of the perturbation.

An additional parameter is the fingertip design (iii) where the notched fingertip has been identified as an optimal solution for achieving form-fit contact and used to achieve a parallel contact plane with the grasped part.

\subsection{Stiffness model}
\label{sec:stiffness}

The stiffness model used in the simulation represents a simplified model of the finger, utilizing the stiffness values in the compliance direction \cite{Hartisch_TMECH}. 
As visualized in Figure \ref{fig:search}, contact is only established at the two contact planes in the fingertip. With the contact and stiffness resulting on the gripped part, sufficient search and assembly forces are provided. For this work it is therefore assumed, that each finger is modeled individually and the stiffness model of the compliant gripper can be sufficiently reduced to the physical contact between the notch of the fingers and the grasped parts utilizing only the diagonal entries of a 6-DOF stiffness matrix.
% The implementation of the model in the dynamic simulation is described later in Sec. \ref{sec:inputs}.

% \begin{figure}[t]
%     \centering
%     \includegraphics[width=0.9\linewidth]{figs/coordinatessim.pdf}
%     \caption{Demonstration of the resulting stiffness model of the compliant fingers with the assembly- and compliance directions}
%     \label{fig:stiffness}
% \end{figure}

\section{Simulation Preparation}
\label{sec:simprep}
This section describes the geometric modelling of the connectors and simulation setup. 
% The dynamic simulation in this works acts as a means for implementing the reduced stiffness model of the compliant grippers in the assembly tasks. To fully implement the assembly tasks, preliminary work is needed, described in the following. 

\subsection{Convex Decomposition}
\begin{figure}[t]
\centering
\scalebox{.9}{ % adjust scale factor as needed
    \begin{tabular}{ccc}
        \subfloat[\label{fig:orig_model}]{
            \includegraphics[width=0.3\columnwidth]{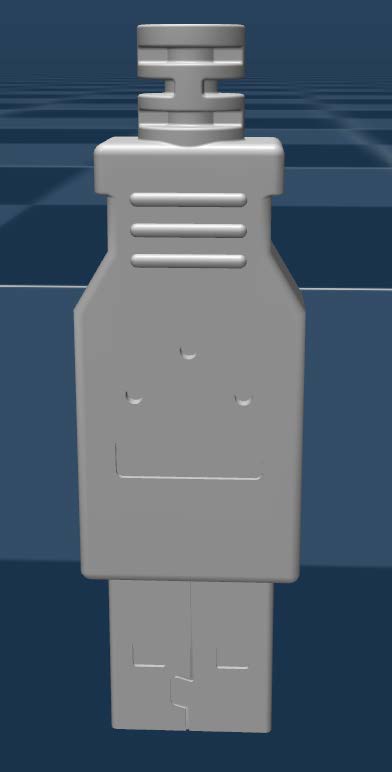}} &
        \subfloat[\label{fig:mujoco_decomp}]{
            \includegraphics[width=0.3\columnwidth]{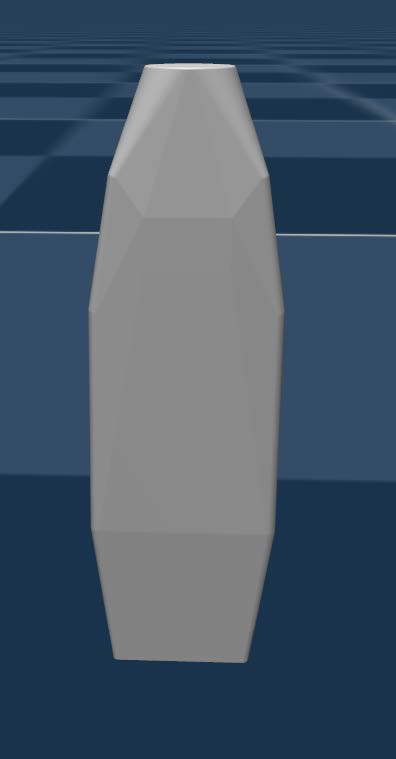}} &
        \subfloat[\label{fig:blender_model}]{
            \includegraphics[width=0.3\columnwidth]{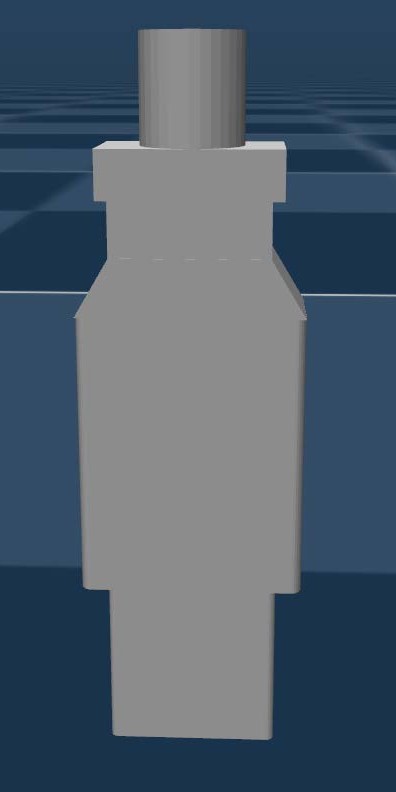}}
    \end{tabular}
}
\caption{Here, the different models used for simulation can be seen: (a) the original model to be simulated, (b) the convex auto-decomposition provided by MuJoCo, and (c) the manual remodeling in Blender.}

\label{fig:decomp}
\end{figure}

% \subsection{Convex Decomposition}
% \begin{figure} [t]
%     \scalebox{.5}
%     \centering
%     \subfloat[
% 	\label{fig:orig_model}]{\includegraphics[width=.25\columnwidth]{figs/orig_model.jpg}}
% 	\hspace{0.2cm}
%     \subfloat[
% 	\label{fig:mujoco_decomp}]{\includegraphics[width=.25\columnwidth]{figs/mujoco_decomp.jpg}}
% 	\hspace{0.2cm}
%     \subfloat[
% 	\label{fig:blender_model}]{\includegraphics[width=.25\columnwidth]{figs/blender_model.jpg}}
%     \caption{here, the different models used for simulation can be seen, in (a) the original model to be simulated is seen, (b) shows the convex auto-decompositioning provided by MuJoCo and (c) visualizes the manual remodeling in Blender}
%      \vspace{-0.5cm}
%     \label{fig:decomp}
% \end{figure}

MuJoCo can only handle convex meshes when performing peg-in-hole operations. Therefore, the connector and socket components, as well as the NIST Task Board itself, need to be decomposed. 
% A convex hull necessitates a straight line between two points in the structure to stay entirely within the structure itself. In contrast, in concave structures straight lines can be outside of the structures, which can be the case, for example, in a bowl shape.

% Because the computation of a collision in MuJoCo must be computed quickly, it only uses convex shapes, where only the points at the surface are of interest. However, with a concave shape, it needs to be additionally checked if a surface point lies inside the structure. 

While MuJoCo offers an in-built option to auto-decompose structures into small convex parts, the approximation for complex structures like a connector, visualized in Fig. \ref{fig:orig_model}, is imprecise, leading to significant changes in tolerance, as seen in Fig. \ref{fig:mujoco_decomp}. Open-source solutions as \cite{wei2022approximate} can be used to decompose the geometries beforehand, providing multiple options to adjust the quality of the decomposition. Alternatively, the model can be decomposed manually using simple convex structures, such as cubes and cylinders in Blender, as demonstrated in Fig. \ref{fig:blender_model}, ensuring a realistic approximation of the real part considering dimensions and tolerances.

\subsection{Robot Movement}

\label{sec:movement}

In MuJoCo the assembly process is carried out by two different movement methods. The tolerance-based function simulates the movement until a certain tolerance between the TCP and the target site's position is reached while the duration-based function simulates the robot's movement for a specified duration. Both functions are based on a similar principle. As long as the condition of the while loop is met, the model is advanced one step forward. Then, the viewer is synchronized to show the updated model. 

The tolerance-based method works by specifying joint positions. By overwriting them, the robot is instantly set to a different position. Utilizing this method, no forces are calculated, as there is no power added to the motors in the joints. This method is used to initially position the robot in the home configuration, and the first part of assembly, where the robot manipulates the connector in preparation of the search-and-insertion movement. This function utilizes the target pose, a set tolerance of $ 1\cdot10^{-3}$, as well as the skipped frames as an argument. The loop stops when the difference of the position of the TCP and the target site drops below the specified tolerance value. 

Subsequently, the search-and-insertion movement results in contact between connector and socket. Therefore, the target position cannot be reached with certainty, making the tolerance-based function unfeasible here. In these phases, the subsequent duration-based method is employed. Here, the robot is continuously moved by changing the actuator values, by adding power to the motors in the joints. It is important to note that the formation and development of contacts can only occur as a result of the forces applied by the actuators and the continuous movement in MuJoCo. Therefore, this method is used for all remaining robot movements involving contacts, i.e. in the contact-rich search-and-insertion phase.

When all positions for the set assembly task are calculated, the simulation can be initiated. The process is identical for all assembly tasks, varying in the starting position and the trajectory distance.
% First, the robot moved to its home position, followed by a movement above the connector within a safe distance. Next, the gripping position was approached and the part grasped, followed by a reverse upwards movement to the previous position. 
For the search-and-insertion movement, the connector is angled at $5-10\degree$, and lowered until a line contact with the outside edge of the socket is established in accordance to the previously defined assembly strategy in \cite{Hartisch_TMECH}. This is visualized in Fig. \ref{fig:search}. After alignment, the connector is moved in the negative x- direction with a slight movement in the z- direction, until the opposite inside plane is touched. Then, the connector is moved back in the positive x- direction until a planar contact with the opposite wall is established. To ascertain a defined position in the socket, the connector is moved in the negative y- direction until contact with the inner side of the socket is established. Now, a no-escapable condition should be established \cite{zhang_compliant_2024}, and demonstrated in Fig. \ref{fig:usb side} and the assembly force can be applied downwards in the z- direction to fully insert the connector.

For both the simulation and the following real-life validation, the maximum tolerable range of each parameter combination is determined from the initial starting position of the search strategy. This position is shifted in compliance direction in $\pm0.1\ mm$ steps until the assembly failed. The tolerable ranges resulted from the maximum shift in both positive and negative directions. 

For the simulation, MuJoCo's built-in viewer is used in combination with a python code. The code loads the model directly from an MJCF and thus gains access to all model parameters and current positions and values so that the simulation can be controlled with it. The implementation of MJCF is described in the following section.

\subsection{Simulation inputs}
\label{sec:inputs}

Models in MuJoCo are built by defining objects, properties, constraints and an environment through a MJCF. The complex mechanical properties of the structured compliant grippers can be modeled by six joints with specific stiffness values. Three slide joints are used for the translational movement and three hinge joints for the rotational movement, therefore defining the direct, diagonal stiffness values of a 6- DOF stiffness matrix, omitting the coupled, off-diagonal stiffness values in correspondance to the stiffness model in Sec. \ref{sec:stiffness}. The three transversal slide joints are positioned in the center of the gripping surface. The placement is only relevant for visualization of the position of the TCP. The actual position of the slide joint has no impact on the gripper's movement as it can only move along the joint's axis. 

The position of the three hinge joints for the rotations about the x-, y- z- axis on the other hand is important as it influences the movement of the gripper and the applied forces due to the leverage. In accordance to the assumptions in \cite{Hartisch_TMECH}, the center of rotation is positioned at the estimated RCC. Here a significant simplification from previously published work on the RCC position has to be noted. As described in \cite{Hartisch_TMECH}, the position of the RCC depends on the infill parameters, i.e. the infill direction which influence the stiffness of the gripper. As only the stiffness values are given as an input, the varying position is not considered in this modeling, instead a fixed RCC position at $21\ mm$ in the z- direction from the contact plane is set.   

% The stiffness values in the compliance- and assembly direction, as indexed in Fig. \ref{fig:schematic}, have been determined from the results of \cite{Hartisch_TMECH}. 
Joints in  MuJoCo have an option to adjust their stiffness and damping values, for which the values obtained in \cite{Hartisch_TMECH} are utilized. As the stiffness values in x- direction and the rotations have not been determined in \cite{Hartisch_TMECH}, assumptions are necessary. Considering the forces while gripping and the leverage, the stiffness is estimated between $0.5 - 2 $ $\frac{Nm}{rad}$. 

For the other joints, the stiffness was set to approximately $10$ times the value of the hinge joint for the x- rotation's values. This was defined according to their real-life behavior, as they behave stiff on the real system. Thus, their impact in the simulation should be close to zero. The damping for every joint is set at least to $1$, which results in critical damping. For bigger values used for the stiffness of the slide joints, the damping values were set iteratively to $100$. 
The stiffness values for each spatial direction and rotation are set at the start of the simulation and then assigned to the corresponding joint. As the simulation is scaled so that $1$ unit equals $1$ meter, the input has to be scaled as well. 
The translational joints values are implemented in $\frac{N}{mm}$, which then gets internally scaled by a factor of $1000$ to match the scale of the simulation. The rotational joint values should be entered in $\frac{N m}{rad}$, which does not require any scaling. An overview of the simulation inputs can be found in Tab. \ref{tab:inputs}.

\begin{table}
    \centering
    \begin{tabular}{|c||c|c|}
    \hline

        Input& Description  & Value \\
        \hline
            \hline
         Tangential Friction & Object and Gripper Interaction & 1 \\
         Tangential Friction & Object and Plug &  0.2\\
         Torsional Friction & - & 0,01 - 0.05 \\
         Solver Impedance &  - & 0.99 \\
         Damping Ratio & Object and Gripper Interaction & 1 \\
         Damping Ratio & Object and Plug & 0.5 \\
         Simulation Timestep & - & $5\cdot10^{-5}$ \\
         Stiffness & Compliance Direction & \cite{Hartisch_TMECH} \\
         Stiffness & Assembly Direction & \cite{Hartisch_TMECH} \\
         Stiffness & z- Rotation & $1 \frac{Nm}{rad}$ \\ 
         Stiffness & x- , y- Rotation & $10 \frac{Nm}{rad}$ \\ 
         Transversal Damping & x- , y-, z- Axis & 1000 \\ 
         Rotational Damping & x- , y-, z- Axis & 1 \\ 
         Grasping Force & - & $5 N$ \\ 
         Gain Factor & Major Axes & 2000 \\
         Gain Factor & Minor Axes & 5000 \\
         Gain Factor & Gripper & 1000 \\
         Tolerance & Tolerance-based function  & $1\cdot10^{-3}$ \\ 
         Integrator & -&  "implicitfast" \\
         Solver & - & Newton\\
         Fixed RCC location & z- Direction & $ 21\ mm$\\ 
        
         \hline
    \end{tabular}
    \caption{Simulation Inputs}

    \label{tab:inputs}
\end{table}

\section{Simulation and In-Situ Experiments}
\label{sec:sims}

The simulation is performed for each NIST assembly task in MuJoCo and in-situ utilizing the aforementioned input values. The evaluation is carried out by observing the simulation process in the MuJoCo viewer. A successful assembly is defined as the full insertion of the connector inside the socket
% The simulation can now be performed for each assembly task. The stiffness values from \cite{Hartisch_TMECH} are utilized as input values and the tolerable range is found as described previously, by shifting in the compliance direction in $\pm0.1\ mm$ steps from the initial starting position of the search strategy, until the assembly fails. When failure occurs, the observed failure case in correspondence to Tab. \ref{tab:failurecasesrev} is noted. Each assembly task is performed in-situ, utilizing the real-life NIST task board.

\begin{figure}[t]
    \centering
    \includegraphics[width=0.8\linewidth]{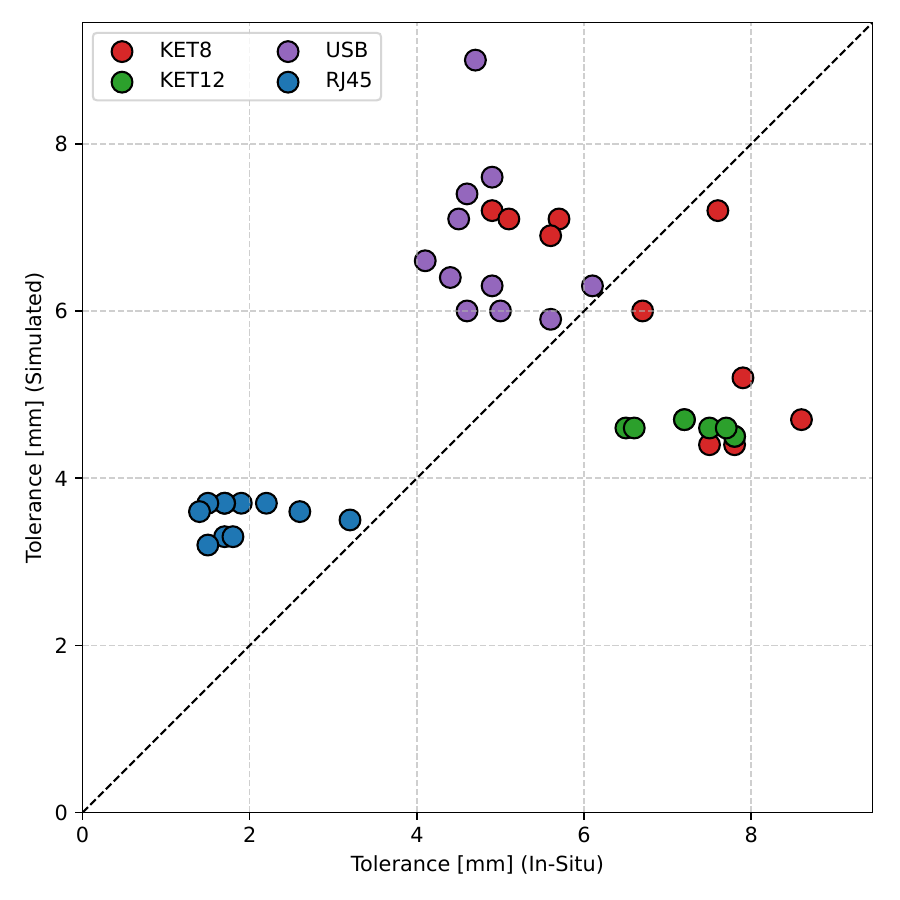}
    \caption{Comparison and Conformity of Simulated and In-Situ Tolerable Ranges}
    \label{fig:comp_tolerance}
\end{figure}

% \begin{figure} [t]
% \scalebox{.5}
%     \centering
%     \subfloat[
% 	\label{fig:comp_tolerance}]{\includegraphics[width=.5\columnwidth]{figs/tolerance_pairs.pdf}}
%     \hfill
%     \subfloat[
% 	\label{fig:degtolKET8}] {\includegraphics[width=.5\columnwidth]{figs/KET8_3.pdf}}
%     \caption{(a) provides a comparison and conformity of simulated and in-situ tolerable ranges. in (b) the simulated (blue) and in-situ (orange) tolerable ranges across the Infill Direction [$\degree$] for the KET8 assembly task are shown}

%     \label{fig:results}
% \end{figure}

\begin{figure} [t]
    \centering
    \includegraphics[width=\columnwidth]{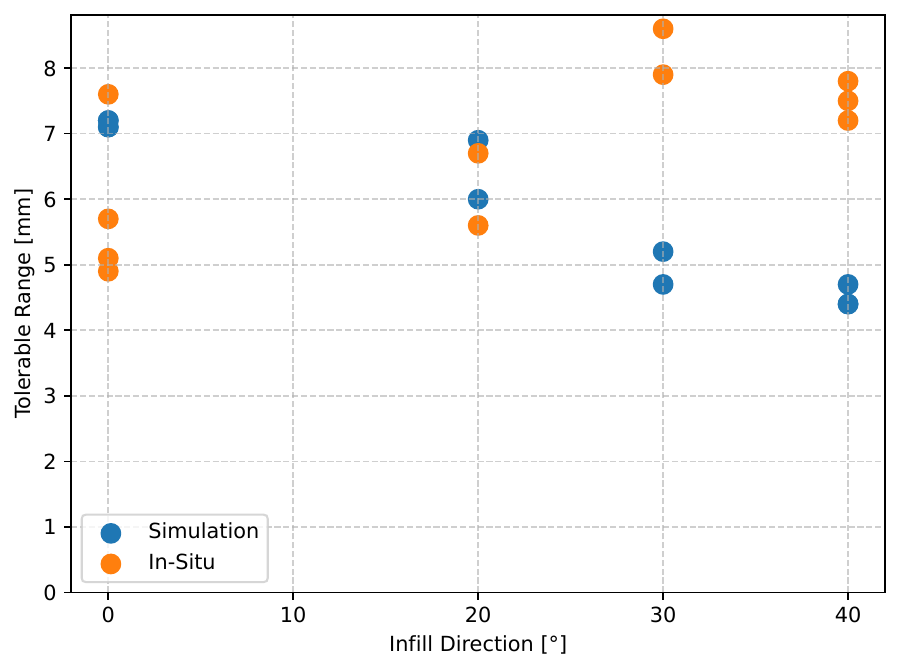}
    \caption{Simulated (blue) and In-Situ (orange) tolerable ranges across the Infill Direction [$\degree$] for the KET8 assembly task}
    \label{fig:degtolKET8}
\end{figure}

\begin{figure} [t]
    \centering
    \includegraphics[width=\linewidth]{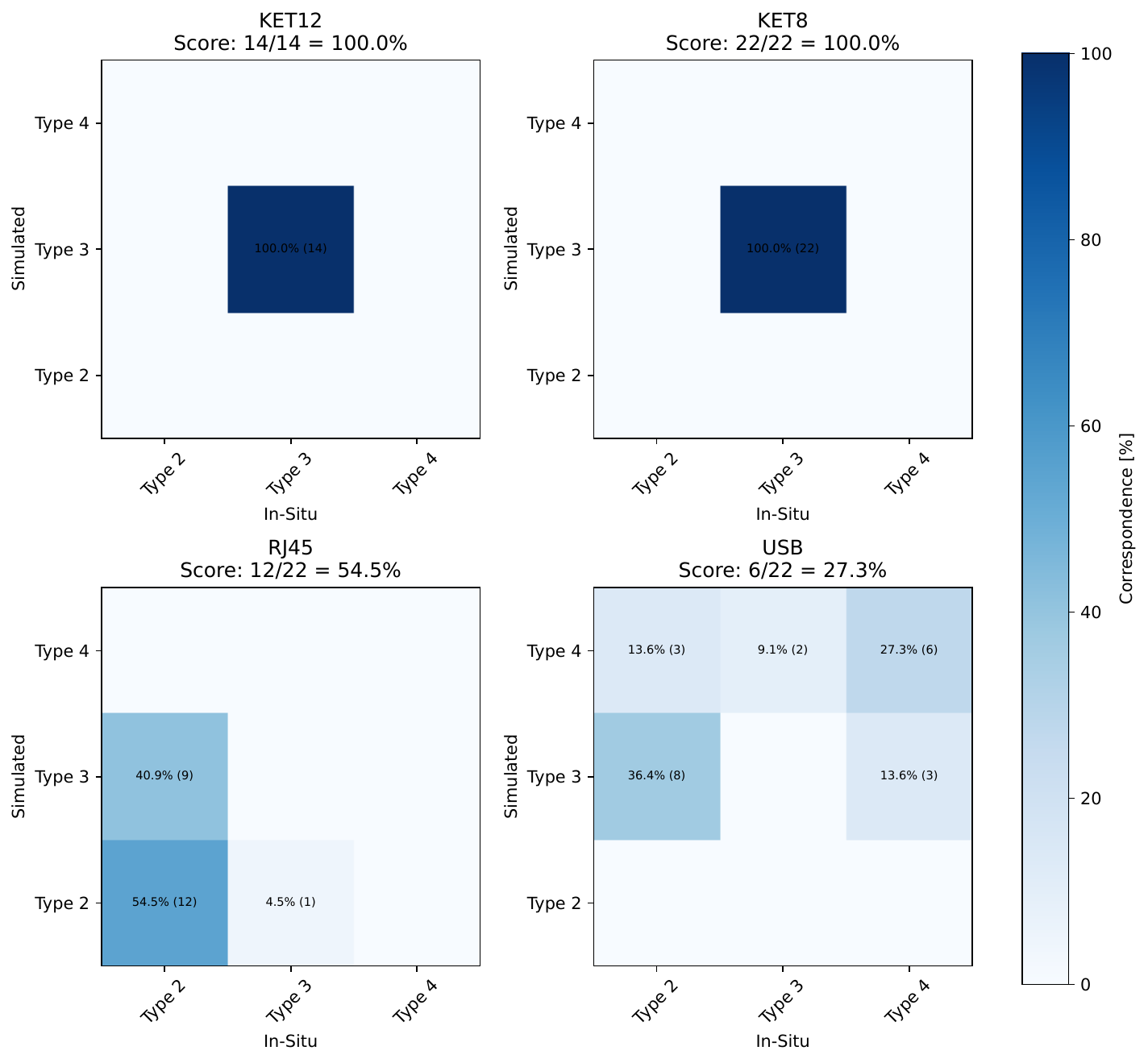}
    \caption{In-Situ (x) and Simulated (y) failure cases observed during the assembly task}
    \label{fig:failurematrix}

\end{figure}

A comparison and conformity of the simulated and in-situ tolerable ranges of the four assembly tasks is given in Fig \ref{fig:comp_tolerance}. 
In the Figures \ref{fig:stifftolKET12}, \ref{fig:stifftolKET8}, \ref{fig:stifftolRJ45} and \ref{fig:stifftolUSB}, the observed tolerable ranges for each stiffness in the compliance direction is shown. The left diagram displays the simulated results and on the right side the corresponding in-situ results are visualized. The infill direction is indexed by five different colors and the increasing infill density with an increasing size of the markers. Subsequently, the assembly tasks are evaluated in real-life, in-situ experiments. 

In Fig. \ref{fig:failurematrix} heatmap visualizations are given of the observed failure cases. The x- axis depicts the failure types of the in-situ trials and the y- axis the observation from the simulated trials. For each evaluation the overall score is given. 

Throughout all in-situ and simulated experiments, failure case 1) has not been detected. While in preliminary experiments in \cite{Hartisch_TMECH} in-finger-slip, i.e. failure case 1) has been noted for low-stiffness-grippers of PETG, these parameter combinations were not used in these experiments. 
% It is assumed that this failure case would be noticeable when the mass of the grasped object exceeds the applicable normal- and friction force of the gripper in the simulation as the failure causes in Tab. \ref{tab:failurecasesrev} indicate. 

The experiments are sorted in an increasing order according to their level of geometric complexity. As the KET components visualized in Figs. \ref{fig:KET12} and \ref{fig:KET8}, consist of a rectangular-shaped peg assembled in a rectangular slot, these are ranked the lowest in terms of complexity. This is followed by the marginally more complex RJ45 task, as additional contact planes in both the plug and socket are introduced here, as seen in Figs. \ref{fig:RJ45}. The task with the highest complexity is the USB assembly task, which is visualized in Figs. \ref{fig:USB}, \ref{fig:missed}, and \ref{fig:loss}. Here, only few marginal contact planes are present on the connector. As illustrated in Fig. \ref{fig:USB} and \ref{fig:missed}, the geometry of the plug adds an additional level of complexity due to the center bar inside the plug.

\subsection{Evaluation of KET12}
\label{sec:val KET12}

\begin{figure} [t]
    \centering
    \includegraphics[width=\columnwidth]{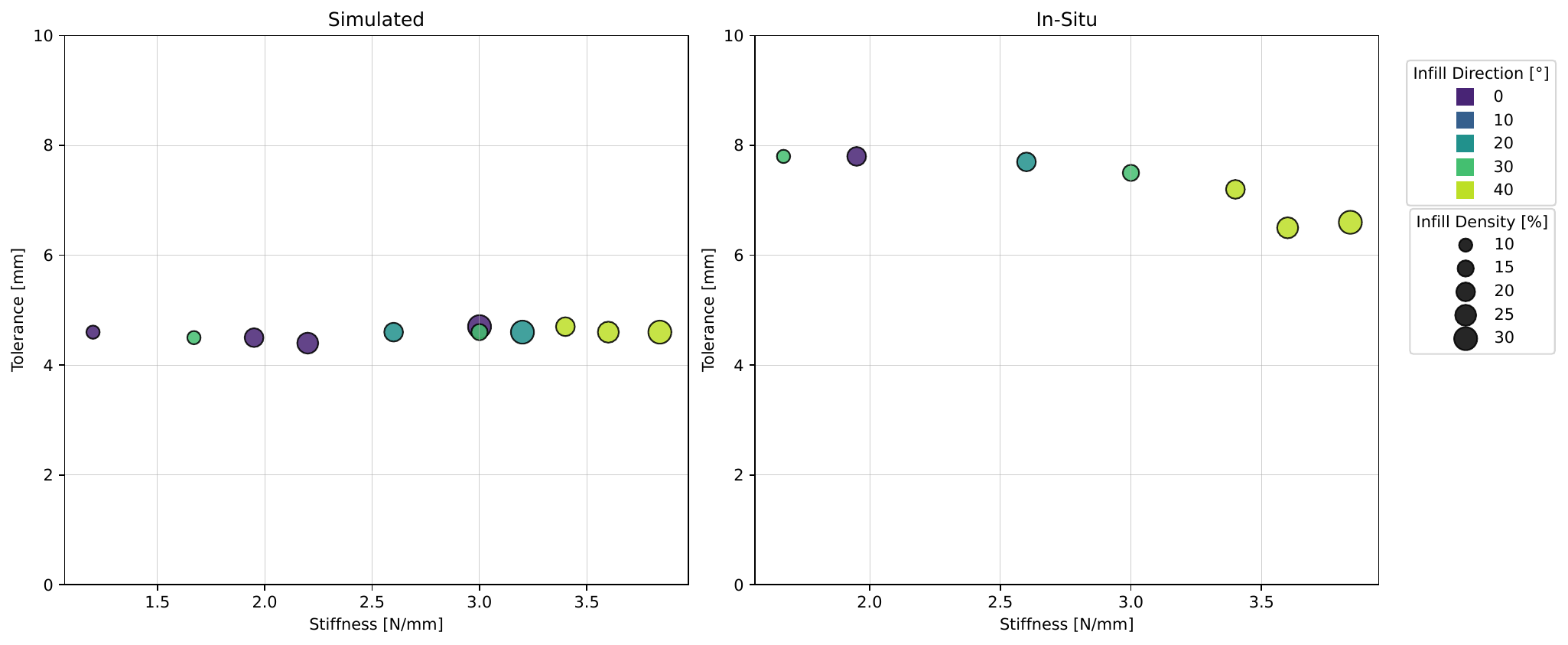}
    \caption{Simulated (left) and In-Situ (right) tolerable ranges across design parameters for the KET12 assembly task}
    \label{fig:stifftolKET12}

\end{figure}

In the KET12 assembly task, the tolerable ranges remain relatively consistent regardless of any changes in the design parameters, as seen in Fig. \ref{fig:comp_tolerance} and Fig. \ref{fig:stifftolKET12}.  
Optimized parameter combinations with low stiffness in the compliance direction of $1.67 - 1.95\ N/mm$ achieve the highest tolerable ranges in the in-situ trials in this task. The optimum increases the tolerable range by a factor of $\approx 1.2$ (from $6.5\ mm$ to $7.8\ mm$).
The main observed failure type lies within the insertion, indicating an excessive inclination of the metal peg as the contributing factor, which results in jamming. Further, the simulated tolerable ranges are lower than in real-life, which achieve up to $\approx 74 \%$ higher tolerable ranges. Jamming has consistently been observed throughout the trials.

\subsection{Evaluation of KET8}
\label{sec:val KET8}

\begin{figure} [t]
    \centering
    \includegraphics[width=\columnwidth]{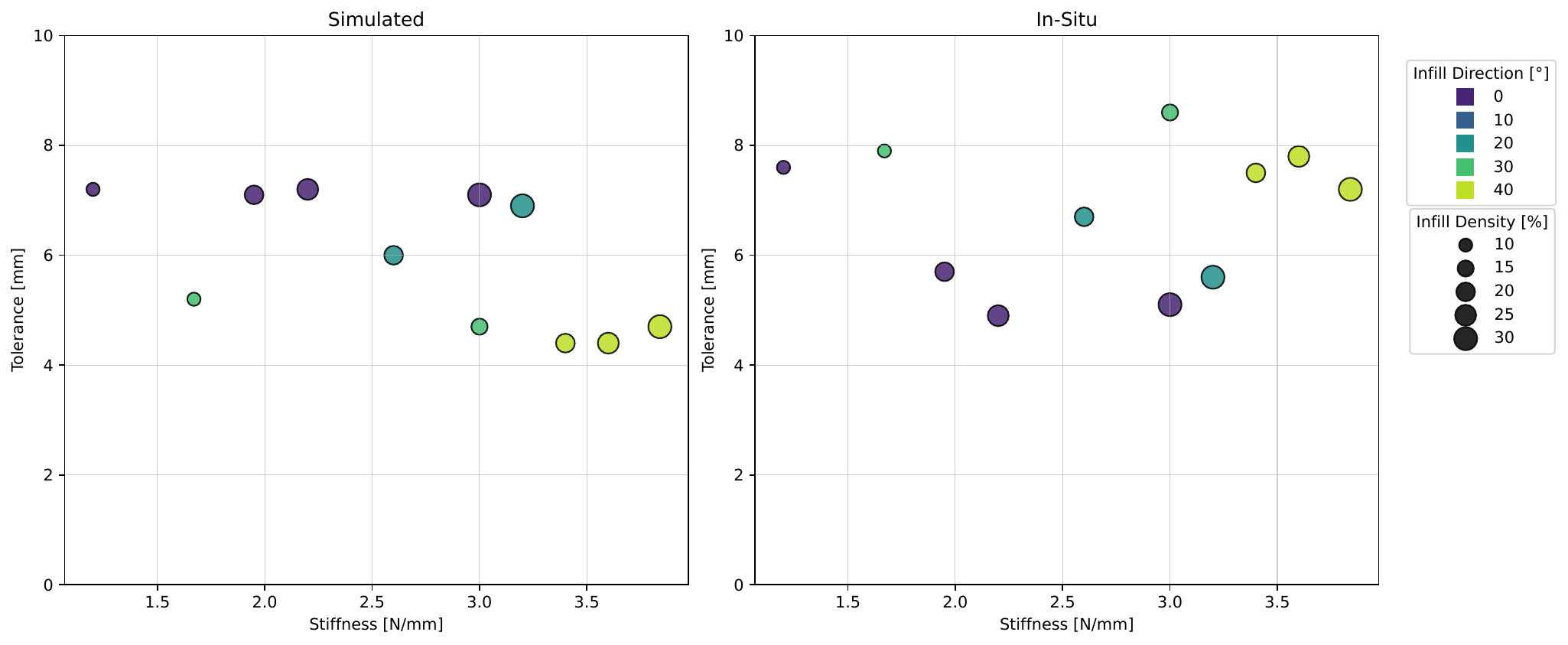}
    \caption{Simulated (left) and In-Situ (right) tolerable ranges across design parameters for the KET8 assembly task}
    \label{fig:stifftolKET8}

\end{figure}

The simulated tolerable ranges for KET8 are highest for low stiffness in compliance direction, of $ \approx7\ mm$, as visualized in Fig. \ref{fig:comp_tolerance} and Fig. \ref{fig:stifftolKET8}. It is found, that the tolerable range decreases up to $2.5\ mm$ with an increasing infill direction. The in-situ experiments, however, achieve a higher range, as seen in Fig. \ref{fig:comp_tolerance} and Fig. \ref{fig:stifftolKET8}. The parameter combinations with $0 \degree$ infill direction achieve the highest range.
As the scattering of the tolerable ranges of the KET8 assembly task has precluded a statement on an observable trend in Fig. \ref{fig:stifftolKET8}, another approach is necessary. As seen in Fig. \ref{fig:degtolKET8}, a trend is observable on the tolerable ranges depending on the infill direction. With an increasing infill direction the tolerable ranges are reduced in the simulation, whereas the real-life validation demonstrates a higher tolerable range with an increased infill direction. It is therefore assumed, that the tolerable ranges of the KET8 task correspond to the effective RCC location influenced through the varying infill direction \cite{Hartisch_TMECH}. The significance of this effect in comparison to the general stiffness is attributed to the slender and long structure of the KET8 beam. It is therefore assumed, that for a correct representation of the real-life behavior for assembly tasks of objects with a high length-to-width ratio a more complex modeling in the simulation needs to be applied which enables the consideration of the varying RCC location. 

The maximum shift in negative and positive direction is limited by connector jamming in both simulation and in-situ trials, as seen in Fig. \ref{fig:failurematrix}, which is denoted by failure type 3). The plug is evaluated as a source for errors, as jamming has been observed due to a rotational movement about the y- axis of the connector when establishing contact with the back plane of the plug. This results in a rotational movement of the object, causing jamming and thus permanently damage of the grippers.  
% In summary, the overall tolerable range is comparable in both trials. Nonetheless, a clear consistency between simulation and in-situ is not given as indicated through the scattering in Fig. \ref{fig:comp_tolerance} and \ref{fig:stifftolKET8}. 
% Regarding the failure types, however, both in-situ and simulation have observed jamming as the only failure case during assembly, demonstrating a high level of consistency. 
Optimized parameter combinations with higher stiffness in the compliance direction of $3\ N/mm$ achieved the highest tolerable ranges in the in-situ trials of this task. The most effective parameter set increased the tolerable range by a factor of $\approx 1.76$ (from $4.9\ mm$ to $8.6\ mm$).

% \begin{figure} [t]
%     \centering
%     \includegraphics[width=\columnwidth]{figs/KET8_2.pdf}
%     \caption{Simulated (blue) and In-Situ (orange) tolerable ranges across the Infill Direction [$\degree$] for the KET8 assembly task}
%     \label{fig:degtolKET8}
% \end{figure}

\subsection{Evaluation of RJ45}
\label{sec:val RJ45}

\begin{figure} [t]
    \centering
    \includegraphics[width=\columnwidth]{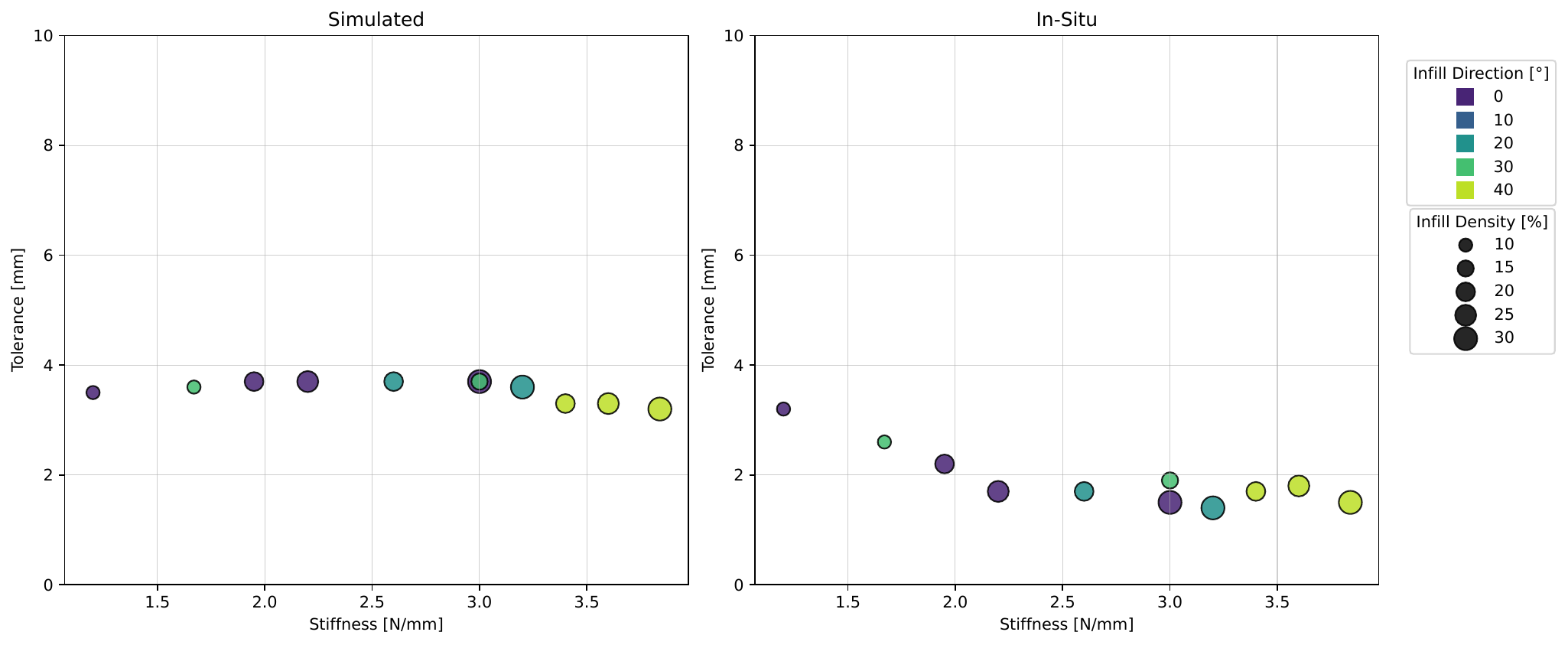}
    \caption{Simulated (left) and In-Situ (right) tolerable ranges across design parameters for the RJ45 assembly task}
    \label{fig:stifftolRJ45}

\end{figure}

For the RJ45 assembly task the overall lowest tolerable range was observed both in the simulation and in-situ experiments as observed in Fig. \ref{fig:comp_tolerance} and \ref{fig:stifftolRJ45}. While the simulated values indicate consistent behavior over stiffness and infill parameter variation, the real-life experiments demonstrate an increasing tolerable range over decreasing infill density values in low-stiffness values of $< 2.5\ N/mm$. 
Optimized parameter combinations with the lowest stiffness in the compliance direction of $1.2\ N/mm$ achieved the highest tolerable ranges in the in-situ trials of this task. Optimal design parameters increased the tolerable range by a factor of $\approx 2.29$ (from $1.4\ mm$ to $3.2\ mm$).
Failure cases 2) and 3) have been noted in the simulated trials, whereas failure case 2) was predominantly observed in the real-life validation, as seen in Fig. \ref{fig:failurematrix}. In both cases the maximum shift was limited by the employed search strategy, which is attributable to the complex geometry of the RJ45 connecting pair. In these cases the connector collided with the outer geometry of the socket during search with an excessive shift in the positive direction causing the connector to miss the socket. This was observed through all trials in the simulated and in-situ experiments, possibly explaining the close similarity of the tolerable ranges and the consistency in the simulation. 
While failure case 3) was observed in the simulation, this was only  noticed once during the in-situ trials, as a failed search strategy has been observed as the main failure case with one exception at the highest tolerable range. This results in a correspondence score of $54.5\ \%$ for the RJ45 assembly task.

\subsection{Evaluation of USB}
\label{sec:val USB}

\begin{figure} [t]
    \centering
    \includegraphics[width=\columnwidth]{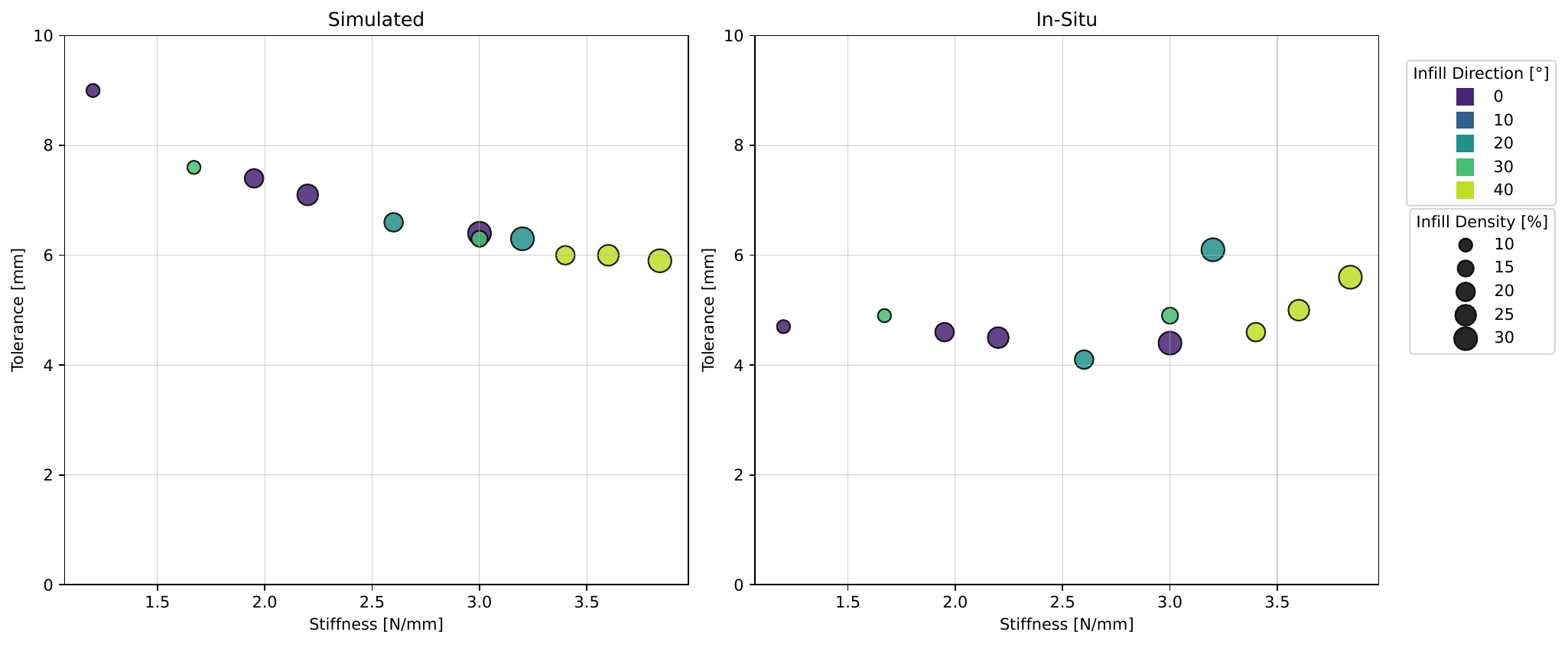}
    \caption{Simulated (left) and In-Situ (right) tolerable ranges across design parameters for the USB assembly task}
    \label{fig:stifftolUSB}

\end{figure}

The simulated USB assembly task demonstrate a clear observable trend on the tolerable ranges in Fig. \ref{fig:comp_tolerance} and \ref{fig:stifftolUSB}. A comparably large tolerance window of up to $9\ mm$ could be achieved, the minimum tolerable range is specified with $5.9\  mm$ for the combination with the highest stiffness. Especially in this trial the differences between simulation and real-life experiments become evident, as the results obtained from the in-situ experiments exhibit a virtually inverse tolerable range over the infill parameters where the highest tolerable range corresponds to the lowest achievable range in the simulated trial. 

Optimized parameter combinations with high stiffness in the compliance direction of $3.2\ N/mm$ achieved the highest tolerable ranges in the in-situ trials of this task. Optimal design parameters increased the tolerable range by a factor of $\approx 1.49$ (from $4.1\ mm$ to $6.1\ mm$).
Additionally, for this assembly scenario different failure types are observed. While jamming and contact loss is determined in the simulation, this does not correspond to the in-situ trials, which are not unambiguously attributable to a specific parameter combination, as summarized in Fig. \ref{fig:failurematrix}. Therefore, the correspondence in this scenario is evaluated with only $\approx 27.3\%$. The following section discusses the assumptions for these deviations.

\section{Conclusion}
\label{subsec:summaryfeasibilityana}

The main motivation behind the implementation of dynamic simulation is the possibility to omit real-life experiments and utilizing simulation tools as a means to anticipate real-life behavior, as contrasted in Fig. \ref{fig:dyn_sim_overview}. 
% As discussed in the introduction, the utilization of dynamic simulation for compliant robot design has not yet been applied, which is why essential preparatory steps have been implemented to achieve necessary convex structures in Sec. \ref{sec:simprep}. The implementation of the robot movement, as well as the reduced compliance model is described in Sec. \ref{sec:movement} and Sec. \ref{sec:inputs}. 

After the preparatory steps in Sec. \ref{sec:simprep}, the failure cases and tolerable ranges in both simulation and in-situ experiments are evaluated in Sec. \ref{sec:sims}. A clear dependency of the tolerable range and the stiffness values is only noticeable for the simulated USB assembly task, whereas the tolerable range remains relatively constant for the simulated RJ45 and KET12 task. This behavior was also partly observable in the in-situ experiments, however an offset of the tolerable ranges, as seen in Fig. \ref{fig:stifftolKET12} and Fig. \ref{fig:stifftolRJ45}, and an almost inverse trend as in Fig. \ref{fig:stifftolUSB} have been noted. 

Especially the slender and long structure of the KET8 assembly task stood out in this work, as initially strong scattering was noticeable in both simulation and in-situ. As described, this is assumed due to an insufficient modeling of the RCC location, which is dependent on the infill direction, as this has allowed for a clearer trend analysis as seen in Fig. \ref{fig:degtolKET8}.  

As a main reason for the discrepancy between simulation and real-life scenarios, insufficient modeling and increased complexity of the tasks are determined, which lead to falsely simulated contact states during assembly. This has been especially noticeable during the USB assembly task. As seen in Fig \ref{fig:search}, the center bar of the USB plug, visualized in Fig. \ref{fig:USB}, is missing, overly simplifying the socket geometry. 
% In addition, it was found that chamfers impaired the search strategy, especially centering, which was noted during the KET8 and KET12 assembly tasks. As chamfers where not included in the CAD models for the simulation but in the real-life components, these have been removed. 
Thus, deviations between the CAD models and real-life parts provided by the NIST task board are critical for the sim2real gap. Regarding the real-life setup, minor differences could mitigate the failure type 3) to a failure type 2), especially due to differences in low friction, the initial angle for the established line contact and deviations in the compliance values. While the dependency of tolerable ranges on the stiffness values have indicated deviations, the overall order of magnitude demonstrates similarities. In both simulated and in-situ trials clearly differentiable tolerable ranges are noticeable between the assembly tasks further demonstrating feasibility to simulate the assembly tasks regardless of exact conformity with the real-life values. 

From the obtained results on the tolerable ranges it is apparent that a universal claim on the optimal design parameters cannot be given. In each task a different parameter combination has achieved the highest tolerable ranges. With an optimized parameter combination an increase by factor $\approx 2.29$ was noted, emphasizing the necessity of task-specific optimization, which will be carried out in the proposed future pipeline. 

% \begin{figure} [t]
%     \scalebox{.75}
%     \centering
%     \subfloat[
% 	\label{fig:usbcontactloss}]{\includegraphics[width=.50\columnwidth]{figs/contactlossusb.png}}
% 	\hfill
%     \subfloat[
% 	\label{fig:constrrj}]{\includegraphics[width=.45\columnwidth]{figs/constaintsrj.jpg}}
%     \caption{In (a) the USB assembly task and the limit of the notched fingertip are shown. In (b) the RJ45 assembly task is demonstrated with the complex geometry of the RJ45 plug impeding the search strategy.}

%     \label{fig:limitationsrl}
% \end{figure}

Regarding the congruence of simulation and real-life validation, the heatmaps in Fig. \ref{fig:failurematrix} have indicate a high level of correspondence for the KET assembly tasks, indicating jamming. Again, an increased complexity and overly simplification of the tasks results in higher deviations, which have been noticed for the RJ45 task, as seen in Fig. \ref{fig:failurematrix}, where an excessive offset has either led to a failed search strategy, as the socket was missed or the connector jamming within the socket. The limiting factor was determined the geometry and increased complexity of the socket has been identified, as seen in Fig. \ref{fig:RJ45}.

The matching score for the USB assembly task is the lowest over all tasks, as $\approx 27.3\ \%$. Both contact loss and jamming during simulated insertion have been detected, whereas the findings in the real-life experiments where mixed. Either a missed plug, jamming and contact loss have been identified in this case, precluding a clear observable trend. 
The ambiguous classification of failure cases is attributable to the complications arising from handling the USB connector as visualized in Fig. \ref{fig:USB}, where the contact established through the form-fit contact from the notch is strongly limited, presumably omitted or falsely approximated in the simulation, which is indicated in Fig. \ref{fig:search}. In the real-life scenario this leads to pose uncertainties of the connector in the grasp, introducing instabiilities when establishing contact with the socket during the search phase. As described prior, this discrepancy is attributed to the insufficient accuracy of the CAD model of the socket.  
% Additionally, jamming in the simulation did not cohere with either contact loss or a failed search strategy in the real-life experiments. 

In summary, dynamic task simulation has demonstrated feasibility to approximate real-life behavior in regards of the tolerable ranges and observed failure cases, further providing distinguishable tolerable ranges corresponding to the task complexity. However, significant deviations are present, which are attributable to insufficient modeling of the CAD models, as well as assumed to the reduced compliance model and the simulation input parameters. It is obvious, that the conformity of the CAD files to the actual counterparts has to be as accurate as possible to reduce the sim2real gap. Complex structures as the connectors and sockets of the RJ45 and USB have been especially sensitive to these simplifications as the deviations have shown, whereas a higher level of conformity was observed for simpler geometries as the KET assembly tasks. Future work will incorporate the dynamic simulation in the proposed design-pipeline to further assess the ability to determine task-level optimized parameter combinations of passive compliant structures. 
% Simple geometries have demonstrated coherency in failure cases and (omitting a fixed offset) tolerable ranges, whereas the sensitivity towards simplifications have demonstrated discrepancies in both tolerable ranges and failure cases. 

% \printbibliography  % If using biblatex
\bibliographystyle{IEEEtran} % If using bibtex
\bibliography{lib.bib} % Make sure .bib is exported for bibtex, not biblatex! 

% Included for arXiv, should be removed for IEEE
\appendix

\iffalse
% \documentclass[letterpaper, 10 pt, conference]{ieeeconf}
\documentclass{article}

%\usepackage{clean}
%\shorttitle{TODO}
\usepackage{textcomp, gensymb}

%\documentclass[conference]{IEEEtran}
%\usepackage{geometry}
%\geometry{top=19.1mm, bottom=19.1mm, left=19.1mm, right=19.1mm}

% Figures
\usepackage{adjustbox}
\usepackage{graphicx}
\usepackage{subfig}
\captionsetup{font=footnotesize}
\graphicspath{{./figs}}
\usepackage{epsfig}

% Functionality 
\usepackage{amsmath,amssymb,amsfonts}
\usepackage{algorithm2e}
\usepackage{xcolor}
\usepackage[colorlinks=true, linkcolor=black, urlcolor=cyan, filecolor=black, citecolor=black]{hyperref}
\usepackage{url}
\usepackage{amsmath,bm}
\usepackage{multirow, makecell}
\usepackage{adjustbox}
\usepackage{comment}
\def\code#1{\texttt{#1}}
\let\labelindent\relax
\usepackage{tikz}
\usetikzlibrary{tikzmark}

\usepackage{float}
\usepackage{enumitem}
\usepackage{booktabs}
% \usepackage[style=ieee]{biblatex} % biblatex
\usepackage{cite} % bibtex, should be used for IEEE

\begin{document}

\fi
\section*{Appendix}

\begin{table}[H]
\caption{Evaluation Table for the simulated KET8 Assembly Task}
\centering
\resizebox{\columnwidth}{!}{
\begin{tabular}{| c | c | c || c | c | c | c | c |}
\hline

 \multirow{3}{*}{\textbf{Nr.}}& \textbf{Infill Density} & \textbf{Infill Direction} & \textbf{Stiffness} & \textbf{Stiffness}  & \textbf{Failure Type}  &  \textbf{Failure Type} & \textbf{Tolerable Range} \\

 % & &  &  &  &  &  &  \\

& &  & z- Axis & y- Axis & neg. Offset &  pos. Offset &  \\
\cline{2-8}

& [\%]  & [\degree] & [N/mm] & [N/mm] & [-] &  [-] & [mm] \\

\hline
\hline
1  & 10 & 0     & 52.40   & 1.20    & 3) & 3) & 7.2 \\\hline
2  & 10 & 30    & 25      & 1.67    & 3) & 3) & 5.2 \\\hline
3  & 15 & 30    & 30      & 3.00    & 3) & 3) & 4.7 \\\hline
4  & 20 & 0     & 65      & 1.95    & 3) & 3) & 7.1 \\\hline
5  & 20 & 20    & 45      & 2.60    & 3) & 3) & 6.0 \\\hline
6  & 20 & 40    & 25      & 3.40    & 3) & 3) & 4.4 \\\hline
7  & 25 & 0     & 70      & 2.20    & 3) & 3) & 7.2 \\\hline
8  & 25 & 40    & 30      & 3.60    & 3) & 3) & 4.4 \\\hline
9  & 30 & 0     & 75      & 3.00    & 3) & 3) & 7.1 \\\hline
10 & 30 & 20    & 55      & 3.20    & 3) & 3) & 6.9 \\\hline
11 & 30 & 40    & 35      & 3.84    & 3) & 3) & 4.7 \\
\hline
\end{tabular}}
\label{tab:tolerable ranges KET8}
\end{table}

\begin{table}[H]
\caption{Evaluation Table for the in-situ KET8 Assembly Task}
\centering
\resizebox{\columnwidth}{!}{
\begin{tabular}{| c | c | c ||  c | c | c | c | c |c |}
\hline

 \multirow{3}{*}{\textbf{Nr.}}& \textbf{Infill Density} & \textbf{Infill Direction}  & \textbf{Stiffness}  & \textbf{Failure Type}  &  \textbf{Failure Type} & \textbf{Tolerable Range} &\textbf{Deviation} & \textbf{Deviation} \\

 % & &  &  &  &  &  &  \\

& &  & y- Axis & neg. Offset &  pos. Offset & & & \\
\cline{2-9}

& [\%]  & [\degree] & [N/mm] & [-] &  [-] & [mm] & [mm] & [\%]\\

\hline
\hline
1  & 10 & 0      & 1.20    & 3) & 3) & 7.6 & 0.4 &5.2 \\\hline
2  & 10 & 30     & 1.67    & 3) & 3) & 7.9 & 2.7&34.1\\\hline
3  & 15 & 30     & 3.00    & 3) & 3) & 8.6 & 3.9 &45.3\\\hline
4  & 20 & 0      & 1.95    & 3) & 3) & 5.7 & 1.4&24.6\\\hline
5  & 20 & 20     & 2.60    & 3) & 3) & 6.7 & 0.7 &10.4\\\hline
6  & 20 & 40     & 3.40    & 3) & 3) & 7.5 & 3.1&41.3\\\hline
7  & 25 & 0      & 2.20    & 3) & 3) & 4.9 & 2.3&47.0\\\hline
8  & 25 & 40      & 3.60    & 3) & 3) & 7.8 & 3.4 & 43.5\\\hline
9  & 30 & 0     & 3.00    & 3) & 3) & 5.1 & 2.0 &39.3\\\hline
10 & 30 & 20     & 3.20    & 3) & 3) & 5.6 & 1.3&23.3\\\hline
11 & 30 & 40    & 3.84    & 3) & 3) & 7.2 & 2.5 &34.7 \\
\hline
\end{tabular}}
\label{tab:rltolerable ranges KET8}
\end{table}

\begin{table} [H]
\caption{Evaluation Table for the simulated KET12 Assembly Task}
\centering
\resizebox{\columnwidth}{!}{
\begin{tabular}{| c | c | c || c | c | c | c | c |}
\hline

 \multirow{3}{*}{\textbf{Nr.}}& \textbf{Infill Density} & \textbf{Infill Direction} & \textbf{Stiffness} & \textbf{Stiffness}  & \textbf{Failure Type}  &  \textbf{Failure Type} & \textbf{Tolerable Range} \\

 % & &  &  &  &  &  &  \\

& &  & z- Axis & y- Axis & neg. Direction &  pos. Direction &  \\
\cline{2-8}

& [\%]  & [\degree] & [N/mm] & [N/mm] & [-] &  [-] & [mm] \\

\hline
\hline
1  & 10 & 0     & 52.40   & 1.20    & 3) & 3) & 4.6 \\\hline
2  & 10 & 30    & 25      & 1.67    & 3) & 3) & 4.5 \\\hline
3  & 15 & 30    & 30      & 3.00    & 3) & 3) & 4.4 \\\hline
4  & 20 & 0     & 65      & 1.95    & 3) & 3) & 4.7 \\\hline
5  & 20 & 20    & 45      & 2.60    & 3) & 3) & 4.6 \\\hline
6  & 20 & 40    & 25      & 3.40    & 3) & 3) & 4.6 \\\hline
7  & 25 & 0     & 70      & 2.20    & 3) & 3) & 4.5 \\\hline
8  & 25 & 40    & 30      & 3.60    & 3) & 3) & 4.6 \\\hline
9  & 30 & 0     & 75      & 3.00    & 3) & 3) & 4.7 \\\hline
10 & 30 & 20    & 55      & 3.20    & 3) & 3) & 4.6 \\\hline
11 & 30 & 40    & 35      & 3.84    & 3) & 3) & 4.6 \\
\hline
\end{tabular}}
\label{tab:tolerable ranges KET12}
\end{table}

\begin{table}[H]
\caption{Evaluation Table for the in-situ KET12 Assembly Task}
\centering
\resizebox{\columnwidth}{!}{
\begin{tabular}{| c | c | c ||  c | c | c | c | c |c |}
\hline

 \multirow{3}{*}{\textbf{Nr.}}& \textbf{Infill Density} & \textbf{Infill Direction}  & \textbf{Stiffness}  & \textbf{Failure Type}  &  \textbf{Failure Type} & \textbf{Tolerable Range} &\textbf{Deviation} & \textbf{Deviation} \\

 % & &  &  &  &  &  &  \\

& &  & y- Axis & neg. Offset &  pos. Offset & & & \\
\cline{2-9}

& [\%]  & [\degree] & [N/mm] & [-] &  [-] & [mm] & [mm] & [\%]\\

\hline
\hline
1  & 10 & 0      & 1.20    & 3) & 3) &  &  &  \\\hline
2  & 10 & 30     & 1.67    & 3) & 3) & 7.8 & 3.3& 42.3 \\\hline
3  & 15 & 30     & 3.00    & 3) & 3) & 7.5 & 3.1 & 41.3\\\hline
4  & 20 & 0      & 1.95    & 3) & 3) & 7.8 & 3.1 & 39.7\\\hline
5  & 20 & 20     & 2.60    & 3) & 3) & 7.7 & 3.1 & 40.2\\\hline
6  & 20 & 40     & 3.40    & 3) & 3) & 7.2 & 2.6 & 36.1\\\hline
7  & 25 & 0      & 2.20    & 3) & 3) &  &   &\\\hline
8  & 25 & 40      & 3.60    & 3) & 3) & 6.5 & 1.9 & 29.2\\\hline
9  & 30 & 0     & 3.00    & 3) & 3) &  &  & \\\hline
10 & 30 & 20     & 3.20    & 3) & 3) &  & & \\\hline
11 & 30 & 40    & 3.84    & 3) & 3) & 6.6 & 2.0 & 30.3 \\
\hline
\end{tabular}}
\label{tab:rltolerable ranges KET12}
\end{table}

\begin{table} [H]
\caption{Evaluation Table for the simulated RJ45 Assembly Task}
\centering
\resizebox{\columnwidth}{!}{
\begin{tabular}{| c | c | c || c | c | c | c | c |}
\hline

 \multirow{3}{*}{\textbf{Nr.}}& \textbf{Infill Density} & \textbf{Infill Direction} & \textbf{Stiffness} & \textbf{Stiffness}  & \textbf{Failure Type}  &  \textbf{Failure Type} & \textbf{Tolerable Range} \\

 % & &  &  &  &  &  &  \\

& &  & z- Axis & y- Axis & neg. Direction &  pos. Direction &  \\
\cline{2-8}

& [\%]  & [\degree] & [N/mm] & [N/mm] & [-] &  [-] & [mm] \\

\hline
\hline
1  & 10 & 0     & 52.40   & 1.20    & 3) & 2) & 3.5 \\\hline
2  & 10 & 30    & 25      & 1.67    & 3) & 2) & 3.6 \\\hline
3  & 15 & 30    & 30      & 3.00    & 3) & 2) & 3.7 \\\hline
4  & 20 & 0     & 65      & 1.95    & 3) & 2) & 3.7 \\\hline
5  & 20 & 20    & 45      & 2.60    & 3) & 2) & 3.7 \\\hline
6  & 20 & 40    & 25      & 3.40    & 2) & 2) & 3.3 \\\hline
7  & 25 & 0     & 70      & 2.20    & 3) & 2) & 3.7 \\\hline
8  & 25 & 40    & 30      & 3.60    & 2) & 2) & 3.3 \\\hline
9  & 30 & 0     & 75      & 3.00    & 3) & 2) & 3.7 \\\hline
10 & 30 & 20    & 55      & 3.20    & 3) & 2) & 3.6 \\\hline
11 & 30 & 40    & 35      & 3.84    & 3) & 2) & 3.2 \\
\hline
\end{tabular}}
\label{tab:Eval RJ45}
\end{table}

\begin{table}[H]
\caption{Evaluation Table for the in-situ RJ45 Assembly Task}
\centering
\resizebox{\columnwidth}{!}{
\begin{tabular}{| c | c | c ||  c | c | c | c | c |c |}
\hline

 \multirow{3}{*}{\textbf{Nr.}}& \textbf{Infill Density} & \textbf{Infill Direction}  & \textbf{Stiffness}  & \textbf{Failure Type}  &  \textbf{Failure Type} & \textbf{Tolerable Range} &\textbf{Deviation} & \textbf{Deviation} \\

 % & &  &  &  &  &  &  \\

& &  & y- Axis & neg. Offset &  pos. Offset & & & \\
\cline{2-9}

& [\%]  & [\degree] & [N/mm] & [-] &  [-] & [mm] & [mm] & [\%]\\

\hline
\hline
1  & 10 & 0      & 1.20    & 2) & 3) &  3.2 & 0.3 & 9.4\\\hline
2  & 10 & 30     & 1.67    & 2) & 2) & 2.6 & 1.0 & 38.5 \\\hline
3  & 15 & 30     & 3.00    & 2) & 2) & 1.9 & 1.8 & 94.8 \\\hline
4  & 20 & 0      & 1.95    & 2) & 2) & 2.2 & 1.5 & 68.2 \\\hline
5  & 20 & 20     & 2.60    & 2) & 2) & 1.7 & 2.0 & 117.7 \\\hline
6  & 20 & 40     & 3.40    & 2) & 2) & 1.7 & 1.6 & 94.2 \\\hline
7  & 25 & 0      & 2.20    & 2) & 2) & 1.7 & 2.0 & 117.7 \\\hline
8  & 25 & 40      & 3.60    & 2) & 2) & 1.8 & 1.5 & 83.4\\\hline
9  & 30 & 0     & 3.00    & 2) & 2) & 1.5 & 2.2 & 146.7 \\\hline
10 & 30 & 20     & 3.20    & 2) & 2) & 1.4 & 2.2 & 157.2 \\\hline
11 & 30 & 40    & 3.84    & 2) & 2) & 1.5 & 1.7 & 113.4 \\
\hline
\end{tabular}}
\label{tab:rltolerable ranges RJ45}
\end{table}

\begin{table} [H]
\caption{Evaluation Table for the simulated USB Assembly Task}
\centering
\resizebox{\columnwidth}{!}{
\begin{tabular}{| c | c | c || c | c | c | c | c |}
\hline

 \multirow{3}{*}{\textbf{Nr.}}& \textbf{Infill Density} & \textbf{Infill Direction} & \textbf{Stiffness} & \textbf{Stiffness}  & \textbf{Failure Type}  &  \textbf{Failure Type} & \textbf{Tolerable Range} \\

 % & &  &  &  &  &  &  \\

& &  & z- Axis & y- Axis & neg. Direction &  pos. Direction &  \\
\cline{2-8}

& [\%]  & [\degree] & [N/mm] & [N/mm] & [-] &  [-] & [mm] \\

\hline
\hline
1  & 10 & 0     & 52.40   & 1.20    & 4) & 3) & 9.0 \\\hline
2  & 10 & 30    & 25      & 1.67    & 4) & 3) & 7.6 \\\hline
3  & 15 & 30    & 30      & 3.00    & 4) & 3) & 6.3 \\\hline
4  & 20 & 0     & 65      & 1.95    & 4) & 3) & 7.4 \\\hline
5  & 20 & 20    & 45      & 2.60    & 4) & 3) & 6.6 \\\hline
6  & 20 & 40    & 25      & 3.40    & 4) & 3) & 6.0 \\\hline
7  & 25 & 0     & 70      & 2.20    & 4) & 3) & 7.1 \\\hline
8  & 25 & 40    & 30      & 3.60    & 4) & 3) & 6.0 \\\hline
9  & 30 & 0     & 75      & 3.00    & 4) & 3) & 6.4 \\\hline
10 & 30 & 20    & 55      & 3.20    & 4) & 3) & 6.3 \\\hline
11 & 30 & 40    & 35      & 3.84    & 4) & 3) & 5.9 \\
\hline
\end{tabular}}
\label{tab:Eval USB}
\end{table}

\begin{table}[H]
\caption{Evaluation Table for the in-situ USB Assembly Task}
\centering
\resizebox{\columnwidth}{!}{
\begin{tabular}{| c | c | c ||  c | c | c | c | c |c |}
\hline

 \multirow{3}{*}{\textbf{Nr.}}& \textbf{Infill Density} & \textbf{Infill Direction}  & \textbf{Stiffness}  & \textbf{Failure Type}  &  \textbf{Failure Type} & \textbf{Tolerable Range} &\textbf{Deviation} & \textbf{Deviation} \\

 % & &  &  &  &  &  &  \\

& &  & y- Axis & neg. Offset &  pos. Offset & & & \\
\cline{2-9}

& [\%]  & [\degree] & [N/mm] & [-] &  [-] & [mm] & [mm] & [\%]\\

\hline
\hline
1  & 10 & 0      & 1.20    & 4) & 4) &  4.7 & 4.3 & 91.5 \\\hline
2  & 10 & 30     & 1.67    & 2) & 4) & 4.9 & 2.7 & 55.2 \\\hline
3  & 15 & 30     & 3.00    & 3) & 4) & 4.9 & 1.4 & 28.6 \\\hline
4  & 20 & 0      & 1.95    & 4) & 2) & 4.6 & 2.8 & 60.9 \\\hline
5  & 20 & 20     & 2.60    & 2) & 2) & 4.1 & 2.5 & 61.0 \\\hline
6  & 20 & 40     & 3.40    & 4) & 2) & 4.6 & 1.4 & 30.5 \\\hline
7  & 25 & 0      & 2.20    & 4) & 2) & 4.5 & 2.6 & 57.8 \\\hline
8  & 25 & 40      & 3.60    & 2) & 2) & 5.0 & 1.0 & 20.0 \\\hline
9  & 30 & 0     & 3.00    & 4) & 2) & 4.4 & 2.0 & 45.5 \\\hline
10 & 30 & 20     & 3.20    & 4) & 2) & 6.1 & 0.2 & 3.3 \\\hline
11 & 30 & 40    & 3.84    & 3) & 2) & 5.6 & 0.3 & 5.4 \\
\hline
\end{tabular}}
\label{tab:rltolerable ranges USB}
\end{table}

\end{document}